\pdfoutput=1
\RequirePackage{fix-cm}
\documentclass[smallextended]{svjour3}       
\usepackage[a4paper,left=2cm,right=2cm,top=2.8cm,bottom=3.3cm]{geometry}

\smartqed  
\usepackage{graphicx}
\journalname{Machine Translation}
\usepackage[utf8]{inputenc}

\usepackage{url}
\usepackage{booktabs}
\usepackage{graphicx}   
\usepackage{natbib}
\usepackage{tikz}
\usepackage{pgf}
\usepackage{subcaption}
\PassOptionsToPackage{hyphens}{url}

\usetikzlibrary{positioning}
\usetikzlibrary{calc}
\usetikzlibrary{arrows}
\usetikzlibrary{decorations.pathmorphing,decorations.markings}
\usetikzlibrary{shapes}
\usetikzlibrary{shapes.arrows}
\usetikzlibrary{patterns}
\usetikzlibrary{fit}

\usepackage{amsmath}
\usepackage{amssymb}
\usepackage{ifthen}
\usepackage{booktabs}   
\usepackage{microtype}  
\usepackage{dsfont}     
\usepackage[multidot]{grffile}  
\usepackage{rotating}
\usepackage{graphicx}
\usepackage{color}

\DeclareMathOperator*{\argmin}{arg\,min}

\newcommand{\prob}{P}
\DeclareMathOperator* \ident {\boldsymbol{1}}

\newcommand{\vb}{\,|\,}
\newcommand{\vect}[1]{\boldsymbol{#1}}

\newcommand{\params}{\boldsymbol{\theta}}
\newcommand{\data}{\boldsymbol{D}}

\DeclareMathOperator*{\noise}{C} 

\newcommand{\bnd}{\mathbin{+\mkern-10mu+}}
\newcommand{\tbnd}{$\bnd$}

\newcommand{\hparams}{\hat{\params}}
\newcommand{\morph}{m}

\newcommand{\morphseq}{\boldsymbol{s}}

\newcommand{\fixme}[2][]{%
    \ifthenelse{\equal{#1}{}}%
               {\textsl{[\textcolor{red}{#2}]}}%
               {\textsl{[#1: \textcolor{red}{#2}]}}%
    }%
\newcommand{\term}[1]{\textbf{\emph{#1}}}
\newcommand{\examp}[2][]{\emph{``#2''}\ifthenelse{\equal{#1}{}}{}{ (#1)}}

\newcommand{\abbr}[2][]{%
    \ifthenelse{\equal{#1}{}}%
               {#2}%
               {#1 (#2)}}%

\usepackage{numprint}
\npdecimalsign{.}
\npthousandsep{\ensuremath{\;}}
\npfourdigitnosep

\newcommand{\cm}{$\checkmark$}

\newcommand{\eng}{\textsc{eng}}
\newcommand{\nob}{\textsc{nob}}
\newcommand{\cze}{\textsc{cze}}
\newcommand{\slo}{\textsc{slo}}
\newcommand{\fin}{\textsc{fin}}
\newcommand{\est}{\textsc{est}}
\newcommand{\swe}{\textsc{swe}}
\newcommand{\dan}{\textsc{dan}}
\newcommand{\sme}{\textsc{sme}}

\newcommand{\vanillabt}{\cm$^\dagger$}

\definecolor{dblue}{HTML}{3034ce}
\definecolor{dgreen}{HTML}{25aa18}

\begin{document}

\title{Transfer learning and subword sampling for asymmetric-resource one-to-many neural translation}
\titlerunning{Transfer learning and subword sampling for asymmetric one-to-many NMT}

\author{Stig-Arne Grönroos        \and
        Sami Virpioja             \and
        Mikko Kurimo
}


\institute{Stig-Arne Grönroos \at
              Aalto University, Department of Signal Processing and Acoustics, Espoo, Finland\\
              Tel.: +358-40-7398282\\
              \email{stig-arne.gronroos@aalto.fi} \\
              ORCiD: 0000-0002-3750-6924
           \and
           Sami Virpioja \at
             University of Helsinki, Department of Digital Humanities, Helsinki, Finland \\
             and Utopia Analytics, Helsinki, Finland \\
             \email{sami.virpioja@helsinki.fi} \\
             ORCiD: 0000-0002-3568-150X
           \and Mikko Kurimo \at
              Aalto University, Department of Signal Processing and Acoustics, Espoo, Finland\\
              \email{mikko.kurimo@aalto.fi}           
}

\date{Received: date / Accepted: date}

\maketitle

\begin{abstract}
  There are several approaches for improving neural machine translation for low-resource languages:
  Monolingual data can be exploited via pretraining or data augmentation;
  Parallel corpora on related language pairs can be used via parameter sharing or transfer learning in multilingual models;
  Subword segmentation and regularization techniques can be applied to ensure high coverage of the vocabulary.
  We review these approaches in the context of an asymmetric-resource one-to-many translation task,
  in which the pair of target languages are related,
  with one being a very low-resource and the other a higher-resource language.
  We test various methods on three artificially restricted translation tasks---English to Estonian (low-resource) and Finnish (high-resource), English to Slovak and Czech, English to Danish and Swedish---and one real-world task, Norwegian to North Sámi and Finnish.
  The experiments show positive effects especially for scheduled multi-task learning, denoising autoencoder, and subword sampling.
\end{abstract}


\section{Introduction}

\abbr[Machine translation]{MT} has become an important application for \abbr[natural language processing]{NLP},
enabling increased access to the wealth of digital information collected on-line,
and new business opportunities in multilingual markets.
\abbr{MT} has made rapid advances
following the adoption of deep neural networks in the last decade,
with variants of the
\abbr[sequence-to-sequence]{seq2seq, \citealp{kalchbrenner2013recurrent,sutskever2014sequence}}
architecture currently holding the state of the art in \abbr[neural machine translation]{NMT}.
However, the recent success has not applied to all languages equally.
Current state-of-the-art methods require very large amounts of data:
Seq2seq methods have been shown to work well in large data scenarios,
but are less effective for low-resource languages.
The rapid digitalization of society
has increased the availability of suitable parallel training corpora,
but the growth has not distributed evenly across languages.

The amount of data needed to reach acceptable quality can also vary based on language characteristics.
Rich, productive morphology leads to a combinatorial explosion in the number of word forms.
Therefore, a larger corpus is required to reach the same coverage of word forms.
Often the two challenges coincide,
with morphologically complex languages that are also relatively low on resources.

Three distinct types of resources may be available for MT training:
parallel data, monolingual data, and data in related languages.
In the low-resource translation setting,
it is primarily the parallel data that is scarce.
Monolingual data is easier to acquire and typically more abundant.
In addition, there may be related languages with much more abundant resources.

\begin{figure}
  \usetikzlibrary{positioning}
\usetikzlibrary{calc}
\usetikzlibrary{arrows}
\usetikzlibrary{decorations.pathmorphing,decorations.markings}
\usetikzlibrary{shapes}
\usetikzlibrary{shapes.arrows}
\usetikzlibrary{patterns}
\usetikzlibrary{fit}


\tikzset{ shorten <>/.style={ shorten >=#1, shorten <=#1 } }


\tikzstyle{hilight} = [draw=red, inner sep=0mm, very thick, rounded corners]
\tikzstyle{pil} = [draw, very thick, ->]
\tikzstyle{link} = [draw, very thick]

\tikzstyle{topic} = [draw, thick, text centered, align=center, minimum height=7ex]
\tikzstyle{used} = [fill=dblue!40]
\tikzstyle{wrap} = [inner sep=0mm]

\resizebox{\textwidth}{!}{%
\begin{tikzpicture}

\node[topic, used                    ] (full) {Full\\sharing\strut};
\node[topic,       right=1mm of full ] (partial) {Partial\\sharing\strut};
\node[topic, used, right=1mm of partial] (curr) {Scheduled\\multi-task\strut};
\node[topic,       right=1mm of curr] (emb) {Embeddings\strut};
\node[topic, used, right=1mm of emb] (presub) {Subnetwork\\ pretraining\strut};
\node[topic,       right=1mm of presub] (lm) {LM fusion\strut};
\node[topic, used, right=1mm of lm] (bt) {Back-\\translation\strut};
\node[topic, used, right=1mm of bt] (ae) {Auto-\\encoder\strut};
\node[topic, used, right=1mm of ae] (subwreg) {Subword\\ regularization\strut};
\node[topic, used, right=1mm of subwreg] (multiseg) {Multilingual\\ segmentation\strut};

\node[wrap, fit=(full)(partial)] (wshare) {};
\node[wrap, fit=(emb)(lm)] (wpre) {};
\node[wrap, fit=(bt)(ae)] (waug) {};
\node[wrap, fit=(subwreg)(multiseg)] (wsubword) {};

\node[topic, used, above=8mm of wshare] (param) {Parameter\\sharing\strut};
\node[topic, used, above=8mm of curr] (para) {Parallel\\transfer\strut};
\node[topic, used, above=8mm of emb] (seq) {Sequential\\transfer\strut};
\node[topic, used, above=8mm of lm] (pre) {Pretraining\strut};
\node[topic, used, above=8mm of waug] (aug) {Dataset\\augmentation\strut};
\node[topic, used, above=8mm of wsubword] (subwseg) {Subword\\segmentation\strut};

\node[wrap, fit=(param)(seq)] (wcross) {};
\node[wrap, fit=(pre)(aug)] (wmono) {};

\node[topic, used, above=8mm of wcross] (cross) {Crosslingual\\transfer\strut};
\node[topic, used, above=8mm of wmono] (mono) {Exploiting\\monolingual\strut};
\node[topic, used, above=8mm of subwseg] (subw) {Vocabulary\\construction\strut};

\node[wrap, fit=(cross)(subw)] (wareas) {};

\node[topic, used, above=8mm of wareas] (mnmt) {Low-resource multilingual NMT\strut};

\draw[link] (mnmt.south) -- (cross.north);
\draw[link] (mnmt.south) -- (mono.north);
\draw[link] (mnmt.south) -- (subw.north);

\draw[link] (cross.south) -- (param.north);
\draw[link] (cross.south) -- (para.north);
\draw[link] (cross.south) -- (seq.north);
\draw[link] (mono.south) -- (pre.north);
\draw[link] (mono.south) -- (aug.north);
\draw[link] (subw.south) -- (subwseg.north);

\draw[link] (param.south) -- (full.north);
\draw[link] (param.south) -- (partial.north);

\draw[link] (para.south) -- (curr.north);
\draw[link] (seq.south) -- (curr.north);

\draw[link] (seq.south) -- (emb.north);
\draw[link] (seq.south) -- (presub.north);

\draw[link] (pre.south) -- (emb.north);
\draw[link] (pre.south) -- (presub.north);
\draw[link] (pre.south) -- (lm.north);

\draw[link] (aug.south) -- (bt.north);
\draw[link] (aug.south) -- (ae.north);
\draw[link] (aug.south) -- (subwreg.north);

\draw[link] (subwseg.south) -- (subwreg.north);
\draw[link] (subwseg.south) -- (multiseg.north);

\end{tikzpicture}%
}
  \caption{Overview of techniques for improving low-resource multilingual NMT. Techniques highlighted with blue are used in this work.}
  \label{fig:overview}
\end{figure}
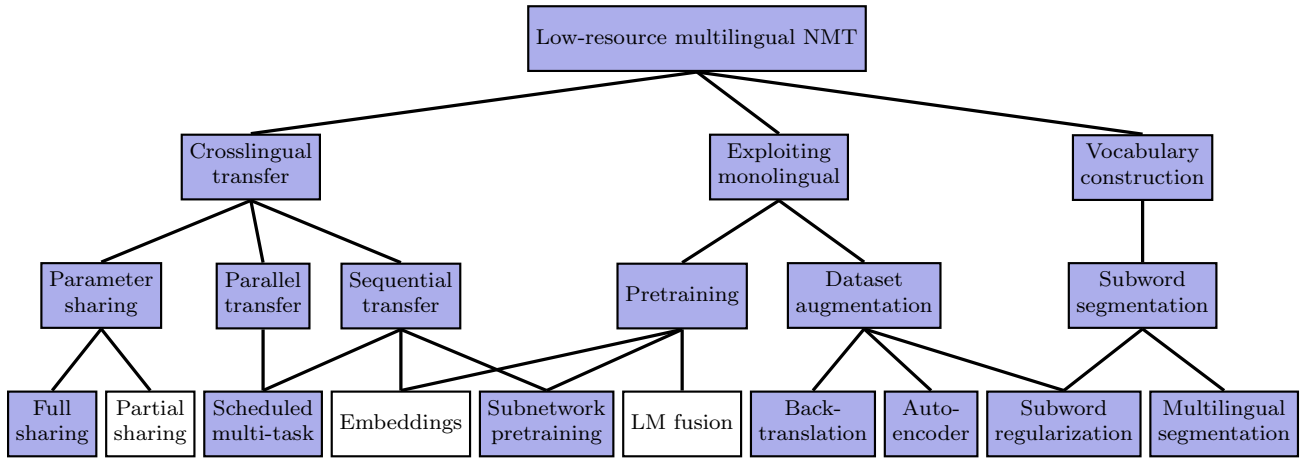

In this work, we consider machine translation \emph{into}
a low-resource morphologically rich language
by means of \term{transfer learning} from a related high-resource target language,
by exploiting available \term{monolingual corpora},
and by exploring the methods and parameters for \term{vocabulary construction}.
Figure~\ref{fig:overview} illustrates an overview of the known techniques for low-resource multilingual NMT; most of them are considered in our experiments.

Our task is a \term{one-to-many} setting in \abbr[multilingual neural machine translation]{MNMT},
as opposed to \term{many-to-one} and \term{many-to-many} settings \citep{luong2016neural}.
As we consider target languages that have different amounts of training resources available,
we call this an \term{asymmetric-resource one-to-many} translation task.
It has three major challenges:

\term{Sparsity.}
Translating into a low-resource is challenging,
especially in the case of a morphologically rich language,
due to a combination of small data and a large target vocabulary.
The resulting data sparsity makes it difficult to estimate statistics
for all but the most frequent items.
Even though continuous-space representations allow neural methods to generalize well,
they learn poorly from low-count events.
Methods like subword segmentation \citep{virpioja2007morphology-aware,sennrich2015neural} can reshape the frequency distribution of the basic units
to reduce sparsity,
and yield a more balanced class distribution in the generator.
Suitable subwords are also beneficial for exploiting transfer from related high-resource languages \citep{gronroos2018cognate},
and from monolingual data.

\term{Data imbalance.}
In multilingual machine translation,
it is very common to have an imbalance between the languages in the training data.
The data can vary in quantity, quality and appropriateness of domain.
Typically all three challenges affect the low-resource languages:
when data is hard to come by, even noisy and out-of-domain data must be used.
The data imbalance is typically addressed by oversampling the low-resource data.
One way to choose the oversampling weights is
using a temperature-based approach to interpolate between
sampling from the true distribution and sampling uniformly~\citep{arivazhagan2019massively}.
An alternative to oversampling the data is to adapt the
gradient scale or learning rate individually for each task~\citep{chen2018gradnorm}.

\term{Task imbalance.}
An NMT system is a conditional language model.
The training signal for the language model is much stronger than
for conditioning on the source.
The conditioning requires training the natural language understanding encoder
and the cross-lingually aligning attention mechanism,
which are both difficult tasks.
High fluency is a known property of NMT~\citep{toral2017multifaceted,koponen2019product}.
When a vanilla \abbr{NMT} system is trained in a low-resource setting,
the learning signal may be sufficient to train the language model,
but insufficient for the conditioning~\citep{ostling2017neural}.
In this case, the MT system degenerates into a fancy language model,
with the output resembling generated nonsense,
with possibly high fluency but little relation to the source text.
As an example, Table~\ref{tab:fancylm} shows an output from an Estonian--English translation system trained from parallel data of only 18k sentence pairs.
\citet{mueller2020analysis} observe this language model overfitting phenomenon
in a massively multilingual but low-resource setting using Bible translations as the corpus.

\begin{table}[t]
\centering
\caption{Example from NMT system overfitted to the language modeling task. \label{tab:fancylm}}
\begin{tabular}{lll}
\toprule
Estonian & Source              & Laktoosi puhul see nii ju ongi! \\
English  & Overfit translation & I've been thinking about it. \\
English  & Reference           & That's the case with lactose! \\
\bottomrule
\end{tabular}
\end{table}

Given these challenges, our research questions include:
\begin{enumerate}
    \item On cross-lingual transfer, is it better to use sequential (pretraining followed by fine-tuning) or parallel (all tasks at the same time) transfer, or something in between?
    \item On exploiting monolingual data:
    \begin{enumerate}
        \item For which languages should one add monolingual auxiliary tasks?
              Is it useful to have a target-language autoencoder in addition to the back-translation strategy, where synthetic training data is generated by a target-to-source translation model?
        \item What kind of noise models are most useful for the denoising sequence autoencoder task?
    \end{enumerate}
    \item On vocabulary construction:
    \begin{enumerate}
        \item What is a suitable granularity of subword segmentation for the low-resource task?
        \item Does it matter what data-driven segmentation method is used?
        \item Does subword regularization (sampling different segmentations for the same word forms) help?
    \end{enumerate}
    \item On available data and languages:
    \begin{enumerate}
        \item When data is very scarce, is it better to train a small model on the low-resource data,
or a larger model using also the auxiliary data?
        \item Is cross-lingual transfer more useful than transfer from monolingual tasks?
        \item How does the amount of the data available for the low-resource language affect the translation quality?
        \item How important is language relatedness for the cross-lingual transfer?
    \end{enumerate}
\end{enumerate}

As methodological contributions for NMT,
we formulate a scheduled multi-task learning technique for asymmetric-resource cross-lingual transfer,
propose our recently introduced Morfessor EM+Prune method \citep{gronroos2020morfessor} for learning the subword vocabulary,
and introduce a taboo sampling task for improving the modeling of segmentation ambiguity.
We include experiments using three diverse language families,
with Estonian, Slovak and Danish as simulated low-resource target languages.
We also contribute a Norwegian bokmål to North Sámi translation system,
the first NMT system for this target language, to the best of our knowledge.

In the next three sections, we will discuss the different techniques for cross-lingual transfer, exploiting monolingual data, and vocabulary construction.
Then we will describe our experimental setup and discuss the results for four different groups of languages, and finally summarize our findings.

\section{Cross-lingual transfer} 

Multilingual training allows exploiting cross-lingual transfer between related languages
by training a single model to translate between multiple language pairs.
This is a form of \term{multi-task learning}~\citep{caruana1998multitask},
in which each language pair in the training data can be seen as a separate learning task
\citep{luong2015multi-task}.
The low-resource language is the main task,
and at least one related high-resource language is used as an auxiliary task.
The cardinality of the multilingual translation has an effect:
cross-lingual transfer is easier in the many-to-one setting
compared to one-to-many \citep{arivazhagan2019massively}.
For a general survey on multilingual translation, see~\citep{dabre2020comprehensive}.

\subsection{Sequential and parallel transfer}

In transfer learning,
knowledge gained while learning one task is transferred to another.
The tasks can either be trained sequentially or in parallel.
Transfer is essential in \term{asymmetric-resource} settings,
in which the amount of training examples for the target task very small,
requiring the learner to rapidly generalize.
\term{Sequential transfer} is a form of adaptation.
In sequential transfer learning,
the \term{pretraining} on a high-resource parent task
is used to initialize and constrain the \term{fine-tuning} training on the low-resource child task.
\citet{zoph2016transfer} apply sequential transfer learning to low-resource neural machine translation.
Sequential transfer carries the risk of
catastrophic forgetting~\citep{mccloskey1989catastrophic,goodfellow2013empirical},
in which the knowledge gained from the first task fades away completely.
Some parameters can be frozen between the two training phases.
This reduces the number of parameters trained from the small data,
which may delay overfitting.

When training tasks in parallel, called \term{multi-task learning},
catastrophic forgetting does not occur.
If the amount of data for different tasks is highly asymmetrical,
careful tuning of the task mixture weights is critical to avoid overfitting on the small task.
Sequential transfer does not require the same tuning, as convergence can be determined for each task separately.

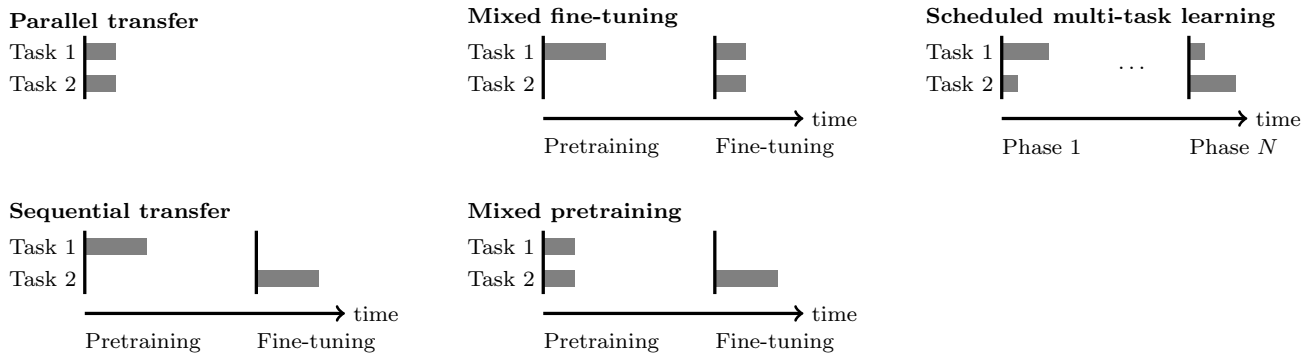
\begin{figure}
  \centering
  \usetikzlibrary{positioning}
\usetikzlibrary{calc}
\usetikzlibrary{arrows}
\usetikzlibrary{decorations.pathmorphing,decorations.markings}
\usetikzlibrary{shapes}
\usetikzlibrary{shapes.arrows}
\usetikzlibrary{patterns}
\usetikzlibrary{fit}


\tikzset{ shorten <>/.style={ shorten >=#1, shorten <=#1 } }


\tikzstyle{hilight} = [draw=red, inner sep=0mm, very thick, rounded corners]
\tikzstyle{pil} = [draw, very thick, ->]
\tikzstyle{distrbar} = [very thick]
\tikzstyle{task} = [inner sep=0mm]
\tikzstyle{phase} = [inner sep=0mm, font=\small]
\tikzstyle{lbl} = [font=\bf, inner sep=0mm]

\resizebox{\textwidth}{!}{%
\begin{tikzpicture}

\node[task] (mtlt1) {Task 1};
\node[task, below=2mm of mtlt1] (mtlt2) {Task 2};
\node[lbl, above=2mm of mtlt1.north west, anchor=south west] (mtl1label) {Parallel transfer};

\coordinate (top) at ($(mtlt1.north east) + (1mm, 1mm)$);
\coordinate (btm) at ($(mtlt2.south east) + (1mm, -1mm)$);
\fill[fill=gray] (top |- mtlt1.north east) rectangle ($(btm |- mtlt1.south east) + (4mm, 0mm)$);
\fill[fill=gray] (top |- mtlt2.north east) rectangle ($(btm |- mtlt2.south east) + (4mm, 0mm)$);
\draw[distrbar] (top) -- (btm);

\node[task, below=19mm of mtlt2] (seqt1) {Task 1};
\node[task, below=2mm of seqt1] (seqt2) {Task 2};
\node[lbl, above=2mm of seqt1.north west, anchor=south west] (seq1label) {Sequential transfer};

\coordinate (top) at ($(seqt1.north east) + (1mm, 1mm)$);
\coordinate (btm) at ($(seqt2.south east) + (1mm, -1mm)$);
\fill[fill=gray] (top |- seqt1.north east) rectangle ($(btm |- seqt1.south east) + (8mm, 0mm)$);
\draw[distrbar] (top) -- (btm);
\node[phase, below=5mm of btm, anchor=north west] (pre) {Pretraining};

\coordinate (top) at ($(top) + (22mm,0)$);
\coordinate (btm) at ($(btm) + (22mm,0)$);
\fill[fill=gray] (top |- seqt2.north east) rectangle ($(btm |- seqt2.south east) + (8mm, 0mm)$);
\draw[distrbar] (top) -- (btm);
\node[phase, below=5mm of btm, anchor=north west] (fine) {Fine-tuning};

\coordinate (timestart) at ($(pre.north west)+(0,2.5mm)$);
\coordinate (timeend)  at ($(fine.north east)+(-4mm,2.5mm)$);
\draw[pil] (timestart) -- (timeend);
\node[at=(timeend), anchor=west, xshift=1mm, phase] (time) {time};

\node[task, right=50mm of mtlt1] (mixedfinet1) {Task 1};
\node[task, below=2mm of mixedfinet1] (mixedfinet2) {Task 2};
\node[lbl, above=2mm of mixedfinet1.north west, anchor=south west] (mixedfine1label) {Mixed fine-tuning};

\coordinate (top) at ($(mixedfinet1.north east) + (1mm, 1mm)$);
\coordinate (btm) at ($(mixedfinet2.south east) + (1mm, -1mm)$);
\fill[fill=gray] (top |- mixedfinet1.north east) rectangle ($(btm |- mixedfinet1.south east) + (8mm, 0mm)$);
\draw[distrbar] (top) -- (btm);
\node[phase, below=5mm of btm, anchor=north west] (pre) {Pretraining};

\coordinate (top) at ($(top) + (22mm,0)$);
\coordinate (btm) at ($(btm) + (22mm,0)$);
\fill[fill=gray] (top |- mixedfinet1.north east) rectangle ($(btm |- mixedfinet1.south east) + (4mm, 0mm)$);
\fill[fill=gray] (top |- mixedfinet2.north east) rectangle ($(btm |- mixedfinet2.south east) + (4mm, 0mm)$);
\draw[distrbar] (top) -- (btm);
\node[phase, below=5mm of btm, anchor=north west] (fine) {Fine-tuning};

\coordinate (timestart) at ($(pre.north west)+(0,2.5mm)$);
\coordinate (timeend)  at ($(fine.north east)+(-4mm,2.5mm)$);
\draw[pil] (timestart) -- (timeend);
\node[at=(timeend), anchor=west, xshift=1mm, phase] (time) {time};

\node[task, below=19mm of mixedfinet2] (mixedpret1) {Task 1};
\node[task, below=2mm of mixedpret1] (mixedpret2) {Task 2};
\node[lbl, above=2mm of mixedpret1.north west, anchor=south west] (mixedpre1label) {Mixed pretraining};

\coordinate (top) at ($(mixedpret1.north east) + (1mm, 1mm)$);
\coordinate (btm) at ($(mixedpret2.south east) + (1mm, -1mm)$);
\fill[fill=gray] (top |- mixedpret1.north east) rectangle ($(btm |- mixedpret1.south east) + (4mm, 0mm)$);
\fill[fill=gray] (top |- mixedpret2.north east) rectangle ($(btm |- mixedpret2.south east) + (4mm, 0mm)$);
\draw[distrbar] (top) -- (btm);
\node[phase, below=5mm of btm, anchor=north west] (pre) {Pretraining};

\coordinate (top) at ($(top) + (22mm,0)$);
\coordinate (btm) at ($(btm) + (22mm,0)$);
\fill[fill=gray] (top |- mixedpret2.north east) rectangle ($(btm |- mixedpret2.south east) + (8mm, 0mm)$);
\draw[distrbar] (top) -- (btm);
\node[phase, below=5mm of btm, anchor=north west] (fine) {Fine-tuning};

\coordinate (timestart) at ($(pre.north west)+(0,2.5mm)$);
\coordinate (timeend)  at ($(fine.north east)+(-4mm,2.5mm)$);
\draw[pil] (timestart) -- (timeend);
\node[at=(timeend), anchor=west, xshift=1mm, phase] (time) {time};

\node[task, right=50mm of mixedfinet1] (scheduledt1) {Task 1};
\node[task, below=2mm of scheduledt1] (scheduledt2) {Task 2};
\node[lbl, above=2mm of scheduledt1.north west, anchor=south west] (scheduled1label) {Scheduled multi-task learning};

\coordinate (top) at ($(scheduledt1.north east) + (1mm, 1mm)$);
\coordinate (btm) at ($(scheduledt2.south east) + (1mm, -1mm)$);
\fill[fill=gray] (top |- scheduledt1.north east) rectangle ($(btm |- scheduledt1.south east) + (6mm, 0mm)$);
\fill[fill=gray] (top |- scheduledt2.north east) rectangle ($(btm |- scheduledt2.south east) + (2mm, 0mm)$);
\draw[distrbar] (top) -- (btm);
\node[phase, below=5mm of btm, anchor=north west] (pre) {Phase 1};

\coordinate (top) at ($(top) + (24mm,0)$);
\coordinate (btm) at ($(btm) + (24mm,0)$);
\fill[fill=gray] (top |- scheduledt1.north east) rectangle ($(btm |- scheduledt1.south east) + (2mm, 0mm)$);
\fill[fill=gray] (top |- scheduledt2.north east) rectangle ($(btm |- scheduledt2.south east) + (6mm, 0mm)$);
\draw[distrbar] (top) -- (btm);
\node[phase, below=5mm of btm, anchor=north west] (fine) {Phase $N$};

\coordinate (dots) at ($(top)!0.5!(btm)$);
\node[phase, left=5mm of dots] (dots) {\textellipsis};

\coordinate (timestart) at ($(pre.north west)+(0,2.5mm)$);
\coordinate (timeend)  at ($(fine.north east)+(-4mm,2.5mm)$);
\draw[pil] (timestart) -- (timeend);
\node[at=(timeend), anchor=west, xshift=1mm, phase] (time) {time};

\end{tikzpicture}%
}
  \caption{Task mixing strategies for transfer learning.
  \label{fig:schedules}}
\end{figure}

It is also possible to combine sequential and parallel transfer.
Figure~\ref{fig:schedules} shows some possible ways of achieving this by mixing the tasks.
One strategy---\term{mixed fine-tuning}---involves first pretraining only on the large task,
and then fine-tuning with a mixture of tasks.
\citet{chu2017empirical} apply this strategy to domain adaptation.
\citet{kocmi2019exploring} try the inverse setting---\term{mixed pretraining}---pretraining on a mixture of tasks and fine-tuning only on the child task.

\citet{kiperwasser2018scheduled} propose generalizing these strategies into \term{scheduled multi-task learning},
in which training examples from different tasks are selected according to a mixing distribution.
The mixing distribution changes during training according to the task-mix schedule.
They experiment with three schedules: constant, exponential and sigmoidal.
We propose a new partwise constant task-mix schedule
suitable for an asymmetric-resource setting with multiple auxiliary tasks.
The task-mix schedule can have an arbitrary number of steps,
any of which can be mixing multiple tasks.
All the other strategies can be recovered by using particular schedules with scheduled multi-task learning.

\subsection{Parameter sharing}

In neural networks, multilingual models are implemented through parameter sharing.
It is possible to share all neural network parameters,
or select a subset for sharing allowing the remaining ones to be language-specific.
Parameter sharing can be either hard or soft.
In hard parameter sharing the exact same parameter matrix is used for several languages.
In soft parameter sharing, each language has its own parameter matrix,
but a dependence is constructed between the corresponding parameters for different languages.

The \term{target language token} \citet{johnson2016googles} and \term{language embedding} \citep{lample2019cross} approaches use hard sharing of all parameters.
In the former, the model architecture is the same as in a language-pair-specific model.
The target language is indicated by a preprocessing step
that prepends to the input a special target language token,
e.g. $\langle$\textsc{to\_fi}$\rangle$ to indicate that the target language is Finnish.
The approach can be scaled to more languages by increasing the capacity of the model,
primarily by increasing the depth in layers \citep{arivazhagan2019massively}.
The latter can be described as a factored representation,
with the language embedding factor
marking the language of each word on the target side.

In contrast to full parameter sharing,
it is also possible to divide the model parameters into shared and
\term{language-specific subnetworks},
e.g. sharing all parameters of the encoder,
while letting each target language have its own decoder.
Parameter sharing can even be controlled on a more fine-grained level~\citep{sachan2018parameter}.
Shared attention~\citep{firat2016multi-way} uses language-specific encoders and decoders
with a shared attention,
while language-specific attention~\citep{blackwood2018multilingual}
does the opposite by sharing only the feedforward sublayers of the decoder,
while using language-specific parameters for the attention mechanisms.

The \term{contextual parameter generator}~\citep{platanios2018contextual}
meta-learns a soft dependency between parameters for different tasks.
It does this by using one neural network (the parameter generator)
to generate from some contextual variables the weights of another network (the model).
\citet{gu2018meta-learning} apply meta-learning
to find initializations that can very rapidly adapt to a new low-resource source language.

\section{Exploiting monolingual data}

While parallel data is the primary type of data used for training MT models,
methods for effectively exploiting the more abundant monolingual data
can greatly increase the number of available examples to learn from.
Use of monolingual data can be viewed as semi-supervised learning:
both labeled (parallel) and unlabeled (monolingual) data are used.
There are two main approaches to exploiting monolingual data in MT:
transfer learning and dataset augmentation.

\subsection{Transfer learning: monolingual pretraining}

In monolingual pretraining, some of the parameters of the final translation model are pretrained on a task 
using monolingual data, possibly using a different loss than the one used during NMT training.
There are several ways to use pretraining:
Pretrain word (or subword) \term{embeddings} for the encoder, decoder, or both.
Pretrain a separate \term{language model} for the target language, and combine it with the predictions of the translation model.
Or, finally, pretrain an entire \term{subnetwork}---encoder or decoder---of the translation model.

\subsubsection{Embeddings}

Source and target embeddings can be pretrained on monolingual data
from the source and target languages, respectively~\citep{digangi2017monolingual}.
Alternatively, joint cross-lingual embeddings can be trained on both~\citep{artetxe2017unsupervised}.
As the embeddings are trained for e.g. a generic contextual prediction task,
this is a form of transfer learning.
The pretrained embeddings can either be frozen or fine-tuned,
by respectively omitting or including them as trainable parameters during NMT training.
\citet{thompson2018freezing} investigate the effects
of freezing various subnetwork parameters---including embeddings---on domain adaptation.
In addition to using monolingual data,
pretrained embeddings can contribute to cross-lingual transfer
in the case of a shared multilingual embedding space~\citep{artetxe2017unsupervised}.
The shared embedding spaces are typically on a word level.

\subsubsection{Language model fusion}

The predictions of a strong language model can be combined with the predictions of the translation model,
either using a separate rescoring step,
or by combining the predictions during decoding, using \term{model fusion}.
This approach is used in statistical machine translation,
where one or more target language models are combined with a statistical translation model.
The approach can also be applied in neural machine translation,
through shallow fusion, deep fusion~\citep{gulcehre2015using},
cold fusion~\citep{sriram2017cold},
or PostNorm~\citep{stahlberg2018simple}.
As a neural machine translation system is already a conditional language model,
it may be preferable to find a way to train the parameters of the NMT system using the monolingual data.

\subsubsection{Subnetwork pretraining}

In subnetwork pretraining, the intent is to pretrain entire network components---the encoder or the decoder---with knowledge about the structure of language.
One way to achieve this using unlabeled data is to apply a language modeling loss during pretraining.
The loss function can either be the traditional next token prediction, or a masked language model.
Alternatively an autoencoder loss can be used.

\citet{domhan2017using}
modify the NMT architecture by adding an auxiliary language model 
loss in the internal layers of the decoder, before attending to the source.
This loss allows the first layers of the decoder to be trained on monolingual data.
They find no benefit of adding the language model loss unless additional monolingual data is used.
Adding monolingual data gives a benefit, but does not outperform back-translation.
\citet{ramachandran2017unsupervised}
pretrain the encoder and decoder with source and target language modeling tasks, respectively.
To prevent overfitting, they use task-mix fine-tuning:
the translation and language modeling objectives
are trained jointly (with equally weighted tasks).
\citet{skorokhodov2018semi}
use both pretraining (on both source and target side) and gated shallow fusion (on the target side) to transfer knowledge from pretrained language models.
Some of the experiments are performed on low-resource data going down to 10k sentence pairs.

\subsection{Dataset augmentation}

The easiest way to improve generalization is to train on more data.
As natural training data is limited,
a practical way to acquire more is to generate additional synthetic data for augmentation.
The main benefit of dataset augmentation is 
as regularization to prevent overfitting to non-robust properties of small data.

Simple ways to generate synthetic data include using a single dummy token on the source side~\citep{sennrich2015improving},
and copying the target to source~\citep{currey2017copied}.
The latter can be interpreted as a target-side autoencoder task without noise.
The largest factor in determining the effectiveness of using synthetic data
is how much the synthetic data deviates from the true data distribution.
To avoid confusing the encoder with synthetic data from a different distribution than the natural data,
it may be beneficial to use a special tag to identify the synthetic data~\citep{caswell2019tagged}.

\subsubsection{Back-translation}

Synthetic data can be self-generated by the model being trained, or a related model.
In machine translation, the best known example of synthetic data is \term{back-translation}~(BT) \citep{sennrich2015improving}.
The process of back-translation begins with the training of a preliminary MT model
in the reverse direction, from target to source.
The target language monolingual data is translated using this model,
producing a synthetic, pseudo-parallel data set with the potentially noisy MT output on the source side.
Because the quality of the translation system used for the back-translation
affects the noisiness of the synthetic data,
the procedure can be improved by iterating
with alternating translation direction~\citep{lample2018phrase}.
\citet{edunov2018understanding} propose adding noise to the back-translation output.
The benefit of noisy back-translation is further analyzed by~\citet{graca2019generalizing},
who recommend turning off label smoothing in the reverse model when combined with sampling decoding.
As a related strategy,
\citet{karakanta2018neural} convert 
parallel data from a high-resource language pair into synthetic data for a related low-resource pair
using transliteration.
\citet{zhang2016exploiting} exploit monolingual data in two ways: through self-learning
by ``forward-translating'' the monolingual source data to create synthetic parallel data,
and by applying a reordering auxiliary task:
the input is the natural source text,
while the output is the source text reordered using rules to match the target word order.

\subsubsection{Subword regularization}

Subword regularization is a technique proposed by \citet{kudo2018subword}
for applying a probabilistic subword segmentation model to generate more variability in the input text.
Each time a word token is used during training,
a new segmentation is sampled for it.
It can be seen as treating the subword segmentation as a latent variable.
While marginalizing over the latent variable exactly is intractable,
the subword regularization procedure approximates it through sampling.

\subsubsection{Denoising sequence autoencoder}

Back-translation is a slow method due to the additional training of the reverse translation model.
A computationally cheaper way to turn monolingual data into synthetic parallel data
is to use a denoising autoencoder as an auxiliary task.
Target language text, corrupted by a noise model, is fed in as a pseudo-source.
Different noise models can be used,
e.g. applying reordering, deletions, or substitutions to the input tokens.
The desired reconstruction output is the original noise-free target language text.

An autoencoder~\citep{bourlard1988auto}
is a neural network that is trained to copy its input to its output.
It applies an encoder mapping from input to a hidden representation, i.e. code $\vect{h} = f(\vect{x})$,
and decoder mapping from code to a reconstruction of the input
$\vect{\hat{x}} = g(\vect{h})$.
To force the autoencoder to extract patterns in the data instead of finding the trivial identity function $\vect{\hat{x}} = \ident(\ident(\vect{x}))$,
the capacity of the code must be restricted somehow.
In the undercomplete autoencoder,
the restriction is in the form of a bottleneck layer with small dimension.
For example, in the original sequence autoencoder~\citep{dai2015semi},
the entire sequence is compressed into a single vector.

In a modern sequence-to-sequence architecture,
the attention mechanism ensures a very large bandwidth between encoder and decoder.
When used as an autoencoder, the network is thus highly overcomplete.
In this case, the capacity of the code has to be controlled by regularization.
Robustness to noise is used as the regularizer
in the \term{denoising autoencoder}~\citep{vincent2008extracting}.
Instead of feeding in the clean example $\vect{x}$,
a corrupted copy of the input is sampled from a noise model $\noise(\vect{\tilde{x}} \vb \vect{x})$.
The denoising autoencoder must then learn to reverse the corruption to reconstruct the clean example.
The use of noise as regularization is a successful technique used e.g. in
Dropout~\citep{srivastava2014dropout},
label smoothing~\citep{szegedy2015rethinking}, and
SwitchOut~\citep{wang2018switchout}.
Also multi-task learning
acts as regularization by claiming some of the capacity of the model.
\citet{belinkov2017synthetic} apply both natural and synthetic noises for NMT evaluation,
finding that standard character-based NMT models are not robust to these types of noise.

There are multiple ways of adding the autoencoder loss to the NMT training.
The simplest one treats the autoencoder task as if it was another language pair for multilingual training,
and involves no changes to the architecture.
When using this type of autoencoder task on target language sentences,
the task cardinality changes into a many-to-one problem:
the model must simultaneously learn a mapping
from source to target and from corrupted target to clean target.
In both tasks the target language is the same.
As the decoder is a conditional language model,
this task strengthens the modeling of the target language.
When using source language sentences,
the model must simultaneously learn a one-to-many mapping
from source to target and from corrupted source to clean source.
Thus the decoder must learn to output both languages.
The task may strengthen the encoder, by increasing its robustness to noise,
and by preventing the encoding from becoming too specific to the target language.
\citet{luong2015multi-task} and \citet{luong2016neural} experiment with various auxiliary tasks,
including this type of autoencoder setup.
They see a benefit of using the autoencoder task, as long as it has a low enough weight in the task mix.
This setup is used also in our experiments.

There are also more complex NMT autoencoder setups.
In \term{dual learning}, the autoencoder is built from source-to-target and target-to-source translation models.
\citet{xia2016dual}
combine source-to-target and target-to-source translations in a closed loop which can be trained jointly,
using two additional language modeling tasks (for source and target respectively),
and reinforcement learning with policy gradient.
\citet{cheng2016semi-supervised} use a dual learning setup to exploit monolingual corpora in both source and target languages.
Their loss consists of four parts: translation likelihoods in both directions, source autoencoder, and target autoencoder.
\citet{tu2016neural} simplify the dual learning setup into
an encoder--decoder--reconstructor network.
The reconstructor attends to the final hidden states of the decoder and thus does not need a separate encoder.
Their aim is to improve adequacy by penalizing undertranslation:
the reconstructor is not able to generate any parts of the sentence omitted by the decoder.

\subsubsection{Noise models for text}

To apply a denoising autoencoder to text, a suitable noise model for text is needed.
In domains such as image and speech, there are very intuitive noises,
including rotating, scaling, and mirroring for images;
and reverberation, time-scale stretching, and pitch shifting for speech.
As text is a sequence of discrete symbols,
where even a small change can have a drastic effect on meaning,
suitable noise models are less intuitive.
It is not feasible to guarantee the noise does not change the correct translation of the input.

\paragraph{Local reordering.}
\citet{lample2017unsupervised} perform a local reordering operation $\sigma$
that they call \term{slightly shuffling} the sentence.
The reordering is achieved by adding to the index $i$ of each token
a random offset drawn from the uniform distribution from 0 to a maximum distance $k$.
The tokens are then sorted according to the offset indices.
This maintains the condition $\forall i \in \{1,n\}, | \sigma(i) - i | \leq k$.

\paragraph{Token deletion.}
Randomly dropping tokens is perhaps the most commonly used noise.
It is the central idea in \term{word dropout}~\citep{iyyer2015deep}.
In word dropout,
each token is dropped according to a Bernoulli distribution parameterized by a tunable dropout probability.

\paragraph{Token insertion.}
Randomly selected tokens can also be inserted into the sentence.
The tokens can be sampled from the entire vocabulary, or from a particular class of tokens.
E.g. \citet{vaibhav2019improving} 
insert three classes of tokens: stop words, expletives, and emoticons.

\paragraph{Token substitution.}
SwitchOut~\citep{wang2018switchout}
applies random substitutions to tokens both in the source and the target sentence.
One benefit of SwitchOut is that it can easily and efficiently
be applied late in the data processing pipeline, even to a numericalized and padded minibatch.
Any noises that affect the length of the sequence are best applied before numericalization.

\paragraph{Token masking.}
Masked language models~\citep{devlin2018bert,song2019mass,lewis2019bart,joshi2020spanbert}
apply a special case of token substitution, randomly substituting tokens or spans of tokens with a mask symbol.

\paragraph{Word boundary noise.}
In a special case of token substitution,
the substituted token is selected deterministically as the token with a word boundary marker
either added or removed.
E.g. \examp{kielinen} would be substituted by \examp{\textvisiblespace kielinen}
and vice versa.
This might improve robustness to compounding mistakes such as \examp[Finnish speaker]{*suomen kielinen}.

\paragraph{Taboo sampling.}
In addition to training the translation model, the idea of subword regularization \citep{kudo2018subword} can be used in the autoencoder.
Here, we propose taboo sampling as a special form of subword regularization for monolingual data.
The method takes a single word sequence as input, and outputs two different segmentations for it.
The two segmentations consist of different subwords, whenever possible.
Only single character morphs are allowed to be reused on the other side, to avoid failure if no alternative exists.
E.g. \examp{unreasonable} could be segmented into \examp{un \tbnd reasonable} on the source side and \examp{unreason \tbnd able} on the target side.
When converted into numerical indices into the lexicon, these two representations are completely different.
The task aims to teach the model to associate with each other the multiple ambiguous ways to segment a word,
by using a segmentation-invariant internal representation.

For each word, one segmentation is sampled in the usual way,
after which another segmentation is sampled using taboo sampling.
During taboo sampling,
all multi-character subwords used in the first segmentation
have their emission probability temporarily set to zero.
To avoid introducing a bias from having all the taboo sampled segmentations on the same side,
the sides are mixed by uniformly sampling a binary mask of the same length as the sentence from the set of masks with half the bits 1.
All words for which the mask bit is set have the source and target segmentations swapped.

\begin{figure}
  \centering
  \usetikzlibrary{positioning}
\usetikzlibrary{calc}
\usetikzlibrary{arrows}
\usetikzlibrary{decorations.pathmorphing,decorations.markings}
\usetikzlibrary{shapes}
\usetikzlibrary{shapes.arrows}
\usetikzlibrary{patterns}
\usetikzlibrary{fit}


\tikzset{ shorten <>/.style={ shorten >=#1, shorten <=#1 } }


\tikzstyle{hilight} = [draw=red, inner sep=0mm, very thick, rounded corners]
\tikzstyle{pil} = [draw, very thick, ->]

\tikzstyle{noise} = [draw, very thick, minimum height=8mm, text width=20mm, align=center]
\tikzstyle{dbl} = [text width=43mm]
\tikzstyle{stochastic} = [fill=dblue!40]

\resizebox{0.75\textwidth}{!}{%
\begin{tikzpicture}

\node[noise, stochastic                      ] (seg1src) {Segment\strut};
\node[noise, stochastic, right=1mm of seg1src] (seg1trg) {Segment\strut};

\node[noise, stochastic, right=8mm of seg1trg] (seg2src) {Segment\strut};
\node[noise, stochastic, right=1mm of seg2src] (seg2trg) {Segment\strut};
\node[noise, stochastic, below=5mm of seg2src] (drop2) {Drop\strut};
\node[noise,             below=5mm of drop2] (pre2) {Target language token\strut};
\node[noise, dbl,        below=5mm of pre2.south west, anchor=north west] (filter2) {Length filter\strut};

\node[noise, dbl, stochastic, right=8mm of seg2trg] (seg3) {Taboo Segment\strut};

\node[noise, stochastic, above=5mm of seg2src] (reorder2) {Reorder\strut};

\node[noise,             at=(seg1src |- pre2)] (pre1) {Target language token\strut};
\node[noise, dbl,        below=5mm of pre1.south west, anchor=north west] (filter1) {Length filter\strut};

\node[noise, dbl,        at=(seg3 |- filter2)] (filter3) {Length filter\strut};
\node[noise,             above=5mm of filter3.north west, anchor=south west] (pre3) {Target language token\strut};

\coordinate (mono) at ($(filter2 |- reorder2)+(0, 10mm)$);
\node[at=(mono), anchor=south] (mono) {Monolingual};
\node[at=(mono -| seg1src)] (src1) {Source};
\node[at=(mono -| seg1trg)] (trg1) {Target};
\node[at=(mono -| seg3)] (monotaboo) {Monolingual};

\coordinate (out) at ($(filter2)+(0, -12mm)$);
\node[at=(out -| seg1src)] (out1src) {Source};
\node[at=(out -| seg1trg)] (out1trg) {Target};

\node[at=(out -| seg2src)] (out2src) {Source};
\node[at=(out -| seg2trg)] (out2trg) {Target};

\node[at=(out -| pre3)] (out3src) {Source};
\node[right=9mm of out3src] (out3trg) {Target};

\draw[pil] (src1) -- (seg1src);
\draw[pil] (trg1) -- (seg1trg);
\draw[pil] (seg1src) -- (pre1);
\draw[pil] (pre1) -- (pre1 |- filter1.north);
\draw[pil] (seg1trg) -- (seg1trg |- filter1.north);
\draw[pil] (out1src |- filter1.south) -- (out1src);
\draw[pil] (out1trg |- filter1.south) -- (out1trg);

\coordinate (monox) at ($(mono.south)+(0,-2mm)$);
\draw[pil] (mono) |- (monox) -| (reorder2.north);
\draw[pil] (mono) |- (monox) -| (seg2trg.north);
\draw[pil] (reorder2) -- (seg2src);
\draw[pil] (seg2src) -- (drop2);
\draw[pil] (drop2) -- (pre2);
\draw[pil] (pre2) -- (pre2 |- filter2.north);
\draw[pil] (seg2trg) -- (seg2trg |- filter2.north);
\draw[pil] (out2src |- filter2.south) -- (out2src);
\draw[pil] (out2trg |- filter2.south) -- (out2trg);

\draw[pil] (monotaboo) -- (seg3);
\draw[pil] (pre3 |- seg3.south) -- (pre3);
\draw[pil] (pre3) -- (pre3 |- filter3.north);
\coordinate (seg3trg) at (out3trg |- seg3.south);
\draw[pil] (seg3trg) -- (seg3trg |- filter3.north);
\draw[pil] (out3src |- filter3.south) -- (out3src);
\draw[pil] (out3trg |- filter3.south) -- (out3trg);

\node[below=11mm of filter1, font=\large] (a) {(a)};
\node[below=11mm of filter2, font=\large] (b) {(b)};
\node[below=11mm of filter3, font=\large] (c) {(c)};

\end{tikzpicture}%
}
  \caption{Transformations applied to data at training time.
           Steps with blue background are part of the stochastic noise model.
           Steps with white background are the deterministic target language token prefixing
           and length filtering.
           Length filtering must be applied after segmentation, which may make the sequence longer.
  \label{fig:noise}}
\end{figure}
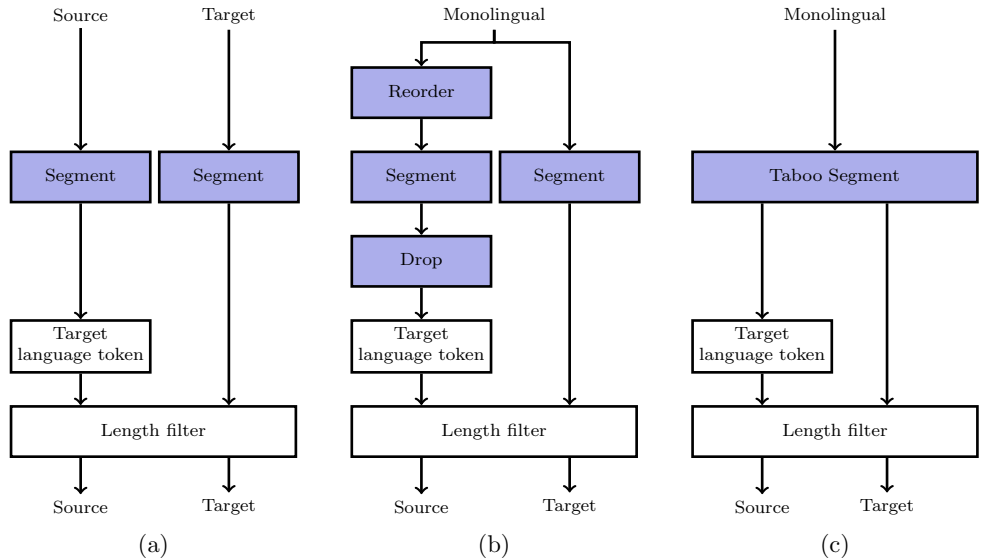

\paragraph{Proposed noise model combinations.}
Our proposed noise model combination is depicted in Figure~\ref{fig:noise}.
It consists of three pipelines:
The pipeline for parallel data (a) consists of only sampling segmentation.
The primary pipeline for monolingual data (b) is a concatenation of multiple noise models:
local reordering, segmentation, and token deletion.
A secondary pipeline for monolingual data (c) uses taboo segmentation.
In all cases the output consists of a pair of source and target sequences.

Observe that the transformations are applied in the data loader at training time,
not as an off-line preprocessing stage.
This allows the noise to be resampled for each parameter update,
which is critical when training continues for multiple epochs of a small dataset.
As a minor downside, the NMT software needs to be modified to accommodate the heavier data loader,
while preprocessing generally requires no modifications to the software.

\section{Vocabulary construction}   

The vocabulary or lexicon of a translation model is the set of basic units or building blocks the text is decomposed into.
In phrase-based machine translation, the standard approach is to use a word lexicon.
Segmentation into subword units has been proposed mostly for morphologically rich languages, for which a word lexicon leads to very high out-of-vocabulary (OOV) rates \citep{lee2004morphological,oflazer2007exploring,virpioja2007morphology-aware},
and character segmentation for closely related languages \citep{tiedemann2009character}.
However, the change of paradigm to neural machine translation has changed also the practice in vocabulary construction:
With the exception of unsupervised translation based on pretrained word embeddings \citep{artetxe2017unsupervised,yang2018unsupervised}, the standard approach for models is segmentation into subword units \citep{sennrich2015neural}.
Some studies aim even to the other extreme, characters \citep{chung2016character-level,costa-jussa2016character} or bytes \citep{costa-jussa2017byte}.

A specific task in subword segmentation is the morphological surface segmentation. There the aim is to split words into morphs, the surface forms of meaning-bearing sub-word units, morphemes.
The concatenation of the morphs is the word, for example
\[ \quad capability \mapsto cap \bnd abil \bnd ity. \]
Unsupervised morphological segmentation, dating back to \citet{harris1955phoneme}, was an active research topic in 2000s and early 2010s \citep{goldsmith2001unsupervised,creutz07acmtslp,hammarstrom2011unsupervised},
and the methods have been evaluated in various NLP applications \citep{kurimo2010morpho,virpioja2011empirical}.
However, in applications based on neural network models, such as NMT, 
the correspondence of the subwords to linguistic morphemes is not of high importance,
as the encoders are able to determine the meaning of the units in context.
Therefore the subword segmentation is typically tuned using other criteria,
such as the size of subword lexicon or the frequency distribution of the units.
Desirable characteristics for a vocabulary to be used in multilingual NMT include:
\begin{enumerate}
    \item \textbf{high coverage} of the training data, without imbalance between languages,
    \item \textbf{a tractable size} for training, and
    \item the right \textbf{level of granularity} for cross-lingual transfer.
\end{enumerate}
Without a high coverage,
some parts of the training data are impossible to represent using the vocabulary.
The unrepresentable parts may be replaced with a special ``unknown'' token.
If the proportion of unknown tokens increases, translation quality deteriorates.
In a multilingual setting, a common approach is to use a shared subword vocabulary between the multiple source or target languages.
In this case, training the segmentation model with a balanced data distribution is important to provide high coverage also for the less resourced languages.

Vocabulary size affects both the memory complexity via the number of network parameters and the computational cost via the length of the sequences and the size of the softmax layer.
When using large vocabularies, e.g. words, the sequences are short, but vocabularies may grow intractably large, particularly for morphologically complex languages.
When using small vocabularies, e.g. characters,
memory requirements are low, but long sequences make training slow, particularly for recurrent networks.

The granularity of the segmentation affects both coverage and size of the lexicon: finer granularity typically means better coverage and smaller lexicon size.
However, within the reasonable limits set by the coverage and size, it is much harder to determine the best possible level of granularity.
Recent research \citep{cherry2018revisiting,kreutzer2018learning,arivazhagan2019massively} indicates that smaller subwords are particularly useful for cross-lingual transfer to low-resource languages in supervised settings.
Exploiting similarity of related languages by increasing the consistency of the segmentation between similar words of the source and target language can also be useful \citep{gronroos2018cognate}.
In unsupervised NMT \citep{artetxe2017unsupervised}, cross-lingual transfer requires basic units to be aligned between languages without use of parallel data.
When starting with pretrained embeddings,
longer units are typically used, as they carry more meaning than short units.
It is therefore an open question how the optimal segmentation granularity varies with the amount of resources available.

Next, we consider different data-driven segmentation methods proposed for machine translation.
This study focuses on segmentation methods applying a \term{unigram language model}.
In the unigram language model,
it is assumed that the morphs in a word occur independently of each other.
Given the parameters $\params$ of the segmentation model,
the probability of a sequence of morphs $\morphseq$
decomposes into the product of the probabilities of the morphs $\morph$ of which it consists:
\begin{equation}
\prob_{\params}(\morphseq) = \prod_{i=1}^{N} \prob_{\params}(\morph_{i}),
\label{eq:unigram}
\end{equation}

\subsection{Byte Pair Encoding}

The most popular method for subword segmentation in the field of NMT is currently the Byte Pair Encoding (BPE) compression algorithm \citep{gage1994new}.
The BPE algorithm iteratively replaces the most frequent pair of bytes in the data with a single unused byte.
In NMT, the algorithm is typically used on characters, and the merging of characters is stopped when the given vocabulary size is reached \citep{sennrich2015neural}.
While BPE is not a probabilistic model, the coding resembles unigram language models in that every subword $\morph_{i}$ is encoded individually. 
As a bottom-up algorithm, BPE is reasonable to use in multilingual settings just by concatenating the corpora before training; this approach is called \term{joint} segmentation \citep{sennrich2015neural}. If the data is balanced over the languages, the frequent words will be constructed in the early steps of the algorithm for all languages.

\subsection{SentencePiece}

SentencePiece \citep{kudo2018subword,kudo2018sentencepiece} is another segmentation method proposed especially for NMT. In contrast to BPE, it defines a proper statistical model for the unigram model in Equation~\ref{eq:unigram}, and tries to find the model parameters that maximize likelihood of the data given a constraint on the vocabulary size.

For training the model, SentencePiece applies the Expectation Maximization (EM) algorithm~\citep{dempster1977maximum}.
The EM algorithm only updates the expected frequencies of the current units; it is not able to add or remove subwords from the vocabulary.
Thus to use EM for the segmentation problem, two other things are needed: a \term{seed lexicon} and a \term{pruning phase}.
The seed lexicon initializes the vocabulary with useful candidate units, and pruning phase removes the least probable units from the model.
Prior to SentencePiece, a similar approach has been proposed by \citet{varjokallio2013learning} for application in automatic speech recognition.

In SentencePiece, the seed lexicon is constructed from the most frequent substrings in the training data.
After initializing the seed lexicon, SentencePiece alternates between the EM phase and the pruning phases until the desired vocabulary size is reached.
In the pruning phase, the subwords are sorted by the reduction in the likelihood function if the subword was removed.
A certain proportion (e.g. 25\%) of the multi-character subwords are pruned at a time, followed by the next EM phase.

\subsection{Morfessor EM+Prune}

Morfessor is a family of generative models for unsupervised morphology induction \citep{creutz07acmtslp}.
Here, consider the Morfessor Baseline method~\citep{creutz2002unsupervised,virpioja2013morfessor} and its recent Morfessor EM+Prune variant \citep{gronroos2020morfessor}.

\subsubsection{Model and cost function}

Morfessor Baseline is applies the unigram language model (Equation~\ref{eq:unigram}).
In contrast to SentencePiece, Morfessor finds 
a point estimate for the model parameters $\hparams$ using
Maximum a Posteriori (MAP) estimation.
The MAP estimate yields a two-part cost function,
consisting of a prior (the lexicon cost) and likelihood (the corpus cost).
The Morfessor prior, inspired by the Minimum Description Length (MDL) principle \citep{rissanen1989stochastic}, favors lexicons containing fewer, shorter morphs.

For tuning the model, \citet{kohonen2010semi-supervised} propose weighting the likelihood with a hyper-parameter $\alpha$:
\begin{equation}
\hparams = \argmin_{\params} \{
    - \log\overbrace{\prob(\params) }^\text{prior} \;
    - \alpha\log\overbrace{\prob(\data \vb \params) }^\text{likelihood} \}
\label{eq:morfessorcost}
\end{equation}
This parameter controls the granularity of the segmentation.
High values increase the weight of each emitted morph in the corpus
(less segmentation),
and low values give a relatively larger weight to a small lexicon
(more segmentation). 

Similar to SentencePiece, Morfessor can be used in subword regularization \citep{kudo2018subword}.
Alternative segmentations can be sampled
from the full data distribution using the forward-filtering backward-sampling algorithm \citep{scott2002bayesian}
or approximatively from an $n$-best list.

\subsubsection{Training algorithm}

The original training algorithm of the Morfessor Baseline method, described in more detail by \citet{creutz2005unsupervised} and \citet{virpioja2013morfessor}, is a local greedy search.
The lexicon is initialized by whole words, and the segmentation proceeds recursively top-down, finding an optimal segmentation into two parts for the current word or subword unit.
Our preliminary studies have indicated that this algorithm does not find as good local optima as the EM algorithm especially for the small lexicons useful in NMT.
As a solution, we have developed a new variant of the method called Morfessor EM+Prune \citep{gronroos2020morfessor}.\footnote{Software available at \url{https://github.com/Waino/morfessor-emprune}.}
It supports the MAP estimation and MDL-based prior of the Baseline model, but implements a new training algorithm based on the EM algorithm and lexicon pruning inspired by SentencePiece.

The training algorithm starts with a seed lexicon and alternates the EM and lexicon pruning steps similarly to SentencePiece.
The prior of the Morfessor model must be slightly modified for use with the EM algorithm,
but the standard prior is used during pruning.
While SentencePiece aims for a predetermined lexicon size,
in Morfessor, the final lexicon size is controlled by the hyper-parameter $\alpha$ (Equation~\ref{eq:morfessorcost}).
To reach a subword lexicon of a predetermined size while using the prior,
Morfessor EM+Prune implements an automatic tuning procedure.
When the estimated change in prior and likelihood are computed separately for each subword,
the value of $\alpha$ that gives exactly the desired size of lexicon after the pruning can be calculated.

In earlier work \citep{gronroos2020morfessor}, we have shown that the EM+Prune algorithm reduces search error during training, resulting in models with lower costs for the optimization criterion.
Moreover, lower costs lead to improved accuracy when segmentation output is compared to linguistic morphological segmentation.
In the present study, we test it for the first time in NMT.

\section{Experiments}

In the experiments, we study how to best exploit
the additional monolingual and cross-lingual resources
for improving machine translation into low-resource morphologically rich languages.
We compare various methods for three major aspects affecting the translation quality:
using cross-lingual transfer, exploiting monolingual data and applying subword segmentation.
The main focus lies on a noise model incorporating the subword segmentation.

We target a one-to-many multilingual setting
with related, morphologically rich languages on the target side.
The related languages include both high- and low-resource languages.
This setting provides a good opportunity for cross-lingual learning,
as the amount of data is highly asymmetric.
Our aim is not to achieve an interlingual representation,
so allowing the encoder to specialize for target languages is acceptable if it improves performance.

\subsection{Data sets}

We perform experiments on four translation tasks,
each consisting of a language triple:
source language (SRC), high-resource target language (HRL) and low-resource target language (LRL).
We only show SRC-LRL translation results, as the goal is to improve this particular translation direction.

The four tasks (LRL in boldface) are:
\begin{enumerate}
    \item English (\eng) to Finnish (\fin) and \textbf{Estonian} (\est),
    \item English to Czech (\cze ) and \textbf{Slovak} (\slo),
    \item English to Swedish (\swe) and \textbf{Danish} (\dan),
    \item Norwegian bokmål (\nob) to Finnish (\fin) and \textbf{North Sámi} (\sme).
\end{enumerate}
In each task the two target languages are related.
The target languages belong to three different language families: Germanic, Balto-Slavic and Uralic.
All target languages are morphologically complex.

We use as parallel corpora
Europarl \citep{koehn2005europarl}, and
OpenSubtitles v2018~\citep{lison2016opensubtitles}, when available.
In addition,
we use the eu, news, and subtitle domains of CzEng v1.7 \citep{czeng2016},
and the UiT freecorpus\footnote{https://victorio.uit.no/freecorpus/}.
The corpora used for each language pair are shown in Table~\ref{tab:corpora}.
The domains for the training data are parliamentary debate, movie subtitles, news and web,
with the exception of North Sámi which contains a mix of many domains.

Our main source of monolingual data is WMT news text%
\footnote{http://www.statmt.org/wmt18/translation-task.html}.
In addition, we use the following monolingual corpora:
skTenTen\footnote{http://hdl.handle.net/11858/00-097C-0000-0001-CCDB-0} and
Categorized News Corpus\footnote{Technical University of Kosice, 2014} for Slovak,
Riksdagens protokoll\footnote{https://spraakbanken.gu.se/eng/resource/rd-prot} for Swedish,
News 2012\footnote{http://hdl.handle.net/11022/0000-0000-2238-B} for Danish,
Aviskorpus\footnote{https://www.nb.no/sprakbanken/show?serial=oai\%3Anb.no\%3Asbr-4\&lang=en} for Norwegian, and
Wikipedia\footnote{sewiki-20191201 dump} for North Sámi.

\begin{table}[h!]
\centering
\caption{Parallel corpora. \label{tab:corpora}}
\begin{tabular}{lllll}
\toprule
     &      & Europarl & OpenSubtitles & Other parallel \\
\midrule
\eng & \cze &          &               & CzEng \\
\eng & \slo & \cm      & \cm           &       \\
\eng & \fin & \cm      & \cm           & Rapid2016, Paracrawl \\
\eng & \est & \cm      & \cm           &       \\
\eng & \swe & \cm      & \cm           &       \\
\eng & \dan & \cm      & \cm           &       \\
\nob & \fin &          & \cm           &       \\
\nob & \sme &          &               & UiT freecorpus \\
\bottomrule

\end{tabular}
\end{table}

For each of the low-resource languages,
we select a subset of 18k sentence pairs.
For \eng-\est, we also perform an experiment where the low-resource subset is repeatedly subsampled down to 3k sentence pairs.
To avoid introducing a domain imbalance in the sampled subset,
the pairs are sampled such that an equal number of sentences 
are selected uniformly at random from each cleaned corpus.
The training data sizes after cleaning and subsampling are shown in Table~\ref{tab:data}.

\begin{table}[ht]
\centering
\caption{Data set sizes after cleaning. \label{tab:data}}
\begin{tabular}{lllrrrrrr}
\toprule
            &      &    &\multicolumn{3}{c}{Parallel} & \multicolumn{3}{c}{Monolingual} \\
\cmidrule(lr){4-6} \cmidrule(lr){7-9}
SRC  & HRL  & LRL  & SRC-HRL & SRC-LRL &   BT  & SRC     & HRL     & LRL   \\
\midrule
\eng & \cze & \slo & 24.7M   & (18k)   &   1M  & 44.3M   & 13.6M   & 27.8M \\
\eng & \fin & \est & 19.4M   & (18k)   &   1M  & 44.3M   &  6.3M   &  3.6M \\
\eng & \swe & \dan & 11.5M   & (18k)   & 750k  & 44.3M   & 10.7M   &  950k \\
\nob & \fin & \sme &  4.9M   & 152k    & 150k  & 40.1M   &  6.3M   &  181k \\
\bottomrule
\end{tabular}
\end{table}

As test sets we use
the WMT newstest2018~\citep{bojar2018findings} for \eng-\est,
the WMT test2011 extended to Slovak by~\citet{galuscakova2011test} for \eng-\slo.
For \eng-\dan\ we use 2k sentence pairs sampled from the JRC-Acquis corpus~\citep{steinberger2006jrc}.
For \nob-\sme\ we use the Apertium story
``Where is James?'', a 48-sentence text with simple language, used as an initial development set for Apertium rule based MT systems~\citep{forcada2011apertium}.

\subsection{Evaluation measures}
When selecting the evaluation measures,
the morphologically rich target languages must be taken into account.
Therefore, we use Character-F$_{1}$~\citep{popovic2015chrf} in addition to BLEU\footnote{mteval-v13a.pl}~\citep{papineni2002bleu}.
To evaluate the performance of systems on rare words,
we use word unigram F$_{1}$ score computed over words occurring less than 5 times in the parallel training data~\citep{sennrich2015neural}.

\subsection{Training details}
\begin{table}[ht]
\centering
\caption{Specifications for the NMT system. \label{tab:hyperparams}}
\begin{tabular}{rlrl}
\toprule
Encoder                     & 8 Transformer layers     & Label smoothing             & 0.1                       \\
Decoder                     & 8 Transformer layers     & Precision                   & 16-bit floating point     \\
Hidden size                 & 1024                     & Minibatch size              & 9200 subword tokens       \\
Filter size                 & 4096                     & Gradient accumulation       & 4 minibatches             \\
Attention heads             & 16                       & Effective minibatch size    & 36800 subword tokens      \\
Adam beta2                  & 0.997                    & Training time               & 100k steps                \\
Warmup                      & noam, 16k steps          & Beam size                   & 8                         \\
Dropout weight              & 0.1                      & Heuristic penalties         & None                      \\
\bottomrule
\end{tabular}
\end{table}

We use the Transformer NMT architecture~\citep{vaswani2017attention}.
Model hyper-parameters are shown in Table~\ref{tab:hyperparams}.
Training takes approximatively 96h on a single V100 GPU, with the data loader in a separate process.
When using scheduled multi-task learning, the mixing distribution is changed after 40k steps.
In all experiments, we apply full parameter sharing using a target language token.
We tune our models towards the best product of the three evaluation measures
(charF$_{1}$, BLEU, rare word F$_{1}$) on a development set.

Back-translation was performed with essentially the same system,
but with sources and targets swapped to achieve a many-to-one configuration.
We mark the back-translation data as synthetic using a special token.

When using subword regularization or denoising autoencoder,
the training data is not simply loaded from disk,
but new random segmentations and noises are sampled each time a training example is used.
To alleviate slowdown, we moved the dataloader and preprocessing pipeline into a separate process,
which communicates the numericalized and padded minibatches to the training process via a multiprocessing queue.
Our data loader is implemented as a fork of OpenNMT-py%
\footnote{Software available at \url{https://github.com/Waino/OpenNMT-py/tree/dynamicdata}.
Later, the dataloader of OpenNMT-py version 2.0 was redesigned to incorporate our proposals.}~\citep{klein2017opennmt}.

With multilingual training, autoencoders and back-translation,
our setting involves a large number of different tasks.
The tasks can be divided by language (HRL, LRL) and by type (translation, autoencoder).
Nearly all runs, with the exception of our vanilla baseline, use a mix of tasks in some or all phases.

\subsection{Results}

In this section, we present the results of ten experiments,
each exploring a separate aspect of asymmetric-resource one-to-many NMT.
We have detailed results for English--Estonian, and verify the central findings on two additional language triples.
Finally, we present some results on the actual low-resource pair Norwegian--North Sámi.

Unless otherwise stated, the compared models are trained using joint Morfessor EM+Prune segmentation with 16k subword vocabulary, cross-lingual scheduled multi-task learning, autoencoder with full noise model, and subword regularization for the translation task.
Our initial results are using autoencoder tasks for all three languages (SRC+HRL+LRL).
Later some of the results were rerun with the better SRC+LRL configuration, which omits the high-resource target language autoencoder.

\subsubsection{Subword segmentation}
\begin{figure}
\centering
  \begin{subfigure}[t]{0.36\textwidth} \centering
  \includegraphics[width=\textwidth]{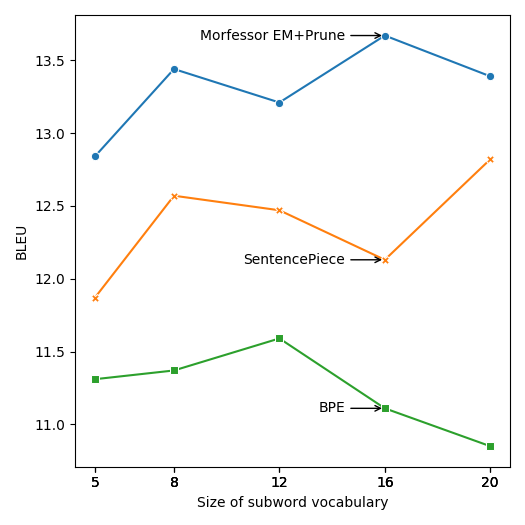}
  \end{subfigure}\begin{subfigure}[t]{0.36\textwidth} \centering
  \includegraphics[width=\textwidth]{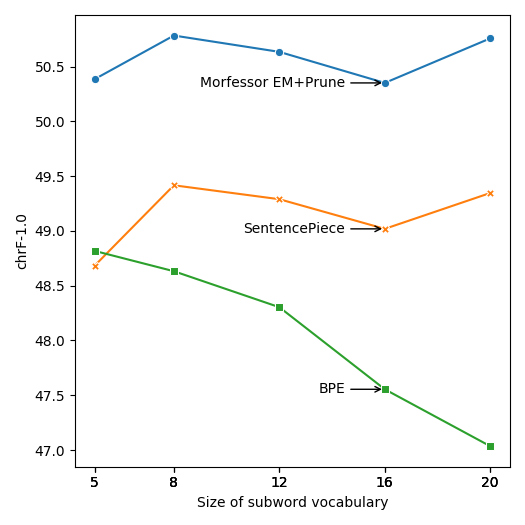}
  \end{subfigure}
  \caption{Varying the subword vocabulary.
           Multilingual models, with SRC+HRL+LRL autoencoder and full noise model,
           except for BPE which are multilingual models without autoencoder or noise.
           Results on English$\rightarrow$Estonian newsdev2018.}
  \label{fig:vocab}
\end{figure}

For subword segmentation, we compare Morfessor EM+Prune to SentencePiece on various vocabulary sizes.
The results are shown in Figure~\ref{fig:vocab}.
There is no clear optimal vocabulary size: in particular for the Character F$_1$ measure the performance remains nearly constant.
On the test set, Morfessor EM+Prune is +0.6 BLEU better than SentencePiece.
The difference is smaller than the +1.48 BLEU difference on the development set, but consistent.
The difference between Morfessor EM+Prune and SentencePiece is similar for the \eng-\dan\ and \eng-\slo\ translation directions.
In preliminary experiments BPE gave 0.65 BLEU worse results than EM+Prune already without subword regularization.
We decided against further experiments using BPE, as it is incompatible with subword regularization.

\subsubsection{Cross-lingual transfer}

Table~\ref{tab:results_multilingual} shows the effect of multilingual training, with and without the autoencoder task.
The cross-lingual transfer from the high-resource language yields the largest single improvement in our experiments.
The multilingual model without autoencoder performs between +10.26 and +12.7 BLEU better than the vanilla model using only LRL parallel data.
Adding an autoencoder loss results in a smaller gain, between +4.97 and +5.55 BLEU.
The gains are partly cumulative for an additional gain of +0.05 to +1.14 BLEU.

\begin{table}[ht]
\centering
\caption{Results for cross-lingual transfer.
         Abbreviations: ML for multilingual, BT for back-translation, AE for autoencoder.\label{tab:results_multilingual}}
\setlength{\tabcolsep}{0.3em}
\begin{tabular}{lllllllrrrrrrrrr}
\toprule
        &          &                     & \multicolumn{3}{c}{Autoencoder} & \multicolumn{3}{c}{\eng--\est} & \multicolumn{3}{c}{\eng--\dan} & \multicolumn{3}{c}{\eng--\slo} \\
\cmidrule(lr){4-6} \cmidrule(lr){7-9} \cmidrule(lr){10-12} \cmidrule(lr){13-15}
Method  & ML       & BT                            &             SRC & HRL & LRL &    chrF1 &  BLEU &  rare & chrF1 &  BLEU &  rare & chrF1 &  BLEU &  rare \\
\midrule
                                  Both &      \cm  &           & \cm &     & \cm &    51.71 & 14.04 & 34.79 & 50.06 & 13.92 & 54.58 & 50.19 & 14.02 & 69.94 \\
                               Only ML &      \cm  &           &     &     &     &    50.09 & 12.90 & 33.20 & 49.57 & 13.13 & 54.21 & 49.83 & 13.97 & 68.79 \\
                              Only AE  &           &           & \cm &     & \cm &    42.65 &  8.19 & 21.59 & 42.26 &  7.60 & 44.48 & 38.97 &  6.25 & 62.51 \\
          Neither \scriptsize(vanilla) &           &           &     &     &     &    29.46 &  2.64 &  6.22 & 31.95 &  2.63 & 30.40 & 23.76 &  1.27 & 36.80 \\
\bottomrule
\end{tabular}
\setlength{\tabcolsep}{0.5em}   
\end{table}

The results for the vanilla model use a smaller configuration, with 4 encoder and 4 decoder layers, and batch size reduced to 2048.
For the vanilla model the small network performed better than the large one,
but when adding either multilingual training or autoencoder, the large network is superior.

\subsubsection{Scheduled multi-task learning}

\begin{figure}
\centering
\begin{minipage}[t]{0.47\textwidth}
  \includegraphics[width=\textwidth]{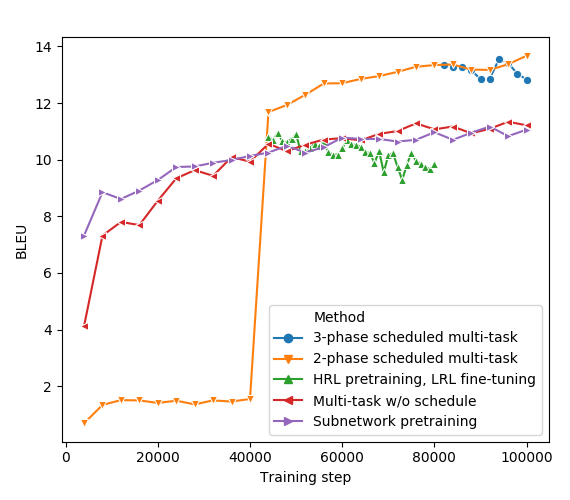}
  \caption{Learning curves on LRL English$\rightarrow$Estonian development set.
           Multilingual models, with SRC+HRL+LRL autoencoder and full noise model.
           Note that up to 40k training steps, the model using scheduled multi-task learning has not seen any LRL data.}
  \label{fig:learningcurve}
\end{minipage}\hfill
\begin{minipage}[t]{0.47\textwidth}
  \usetikzlibrary{positioning}
\usetikzlibrary{calc}
\usetikzlibrary{arrows}
\usetikzlibrary{decorations.pathmorphing,decorations.markings}
\usetikzlibrary{shapes}
\usetikzlibrary{shapes.arrows}
\usetikzlibrary{patterns}
\usetikzlibrary{fit}


\tikzset{ shorten <>/.style={ shorten >=#1, shorten <=#1 } }


\tikzstyle{hilight} = [draw=red, inner sep=0mm, very thick, rounded corners]
\tikzstyle{pil} = [draw, very thick, ->]
\tikzstyle{distrbar} = [very thick]
\tikzstyle{task} = [inner sep=0mm, text width=15mm] 
\tikzstyle{phase} = [inner sep=0mm, font=\small]
\tikzstyle{lbl} = [font=\bf, inner sep=0mm]
\tikzstyle{num} = [font=\tiny, inner sep=0mm]

\resizebox{\textwidth}{!}{%
\begin{tikzpicture}[scale=0.7]

\node[task] (threephaset1) {SRC--HRL};
\node[task, below=2mm of threephaset1] (threephaset2) {SRC--LRL};
\node[task, below=2mm of threephaset2] (threephaset3) {SRC AE};
\node[task, below=2mm of threephaset3] (threephaset4) {HRL AE};
\node[task, below=2mm of threephaset4] (threephaset5) {LRL AE};
\node[lbl, above=2mm of threephaset1.north west, anchor=south west] (threephase1label) {3-phase scheduled multi-task, SRC+HRL+LRL AE};

\coordinate (top) at ($(threephaset1.north east) + (1mm, 1mm)$);
\coordinate (btm) at ($(threephaset5.south east) + (1mm, -1mm)$);
\fill[fill=gray] (top |- threephaset1.north east) rectangle ($(btm |- threephaset1.south east) + (3*9.2mm, 0mm)$);
\coordinate (num) at (top |- threephaset1.north east);
\node[num, white, below right=0.5mm of num, xshift=8mm] () {92};
\coordinate (num) at (top |- threephaset2.north east);
\node[num, below right=0.5mm of num, xshift=8mm] () {0};
\fill[fill=gray] (top |- threephaset3.north east) rectangle ($(btm |- threephaset3.south east) + (3*0.5mm, 0mm)$);
\coordinate (num) at (top |- threephaset3.north east);
\node[num, below right=0.5mm of num, xshift=8mm] () {5};
\coordinate (num) at (top |- threephaset4.north east);
\fill[fill=gray] (top |- threephaset4.north east) rectangle ($(btm |- threephaset4.south east) + (3*0.3mm, 0mm)$);
\node[num, below right=0.5mm of num, xshift=8mm] () {3};
\coordinate (num) at (top |- threephaset5.north east);
\node[num, below right=0.5mm of num, xshift=8mm] () {0};
\draw[distrbar] (top) -- (btm);
\node[phase, below=5mm of btm, anchor=north west] (pre) {Pretraining};

\coordinate (top) at ($(top) + (35mm,0)$);
\coordinate (btm) at ($(btm) + (35mm,0)$);
\fill[fill=gray] (top |- threephaset1.north east) rectangle ($(btm |- threephaset1.south east) + (3*6.7mm, 0mm)$);
\coordinate (num) at (top |- threephaset1.north east);
\node[num, white, below right=0.5mm of num, xshift=8mm] () {67};
\fill[fill=gray] (top |- threephaset2.north east) rectangle ($(btm |- threephaset2.south east) + (3*2.2mm, 0mm)$);
\coordinate (num) at (top |- threephaset2.north east);
\node[num, below right=0.5mm of num, xshift=8mm] () {22};
\coordinate (num) at (top |- threephaset3.north east);
\node[num, below right=0.5mm of num, xshift=8mm] () {0};
\coordinate (num) at (top |- threephaset4.north east);
\node[num, below right=0.5mm of num, xshift=8mm] () {0};
\fill[fill=gray] (top |- threephaset5.north east) rectangle ($(btm |- threephaset5.south east) + (3*1.1mm, 0mm)$);
\coordinate (num) at (top |- threephaset5.north east);
\node[num, below right=0.5mm of num, xshift=8mm] () {11};
\draw[distrbar] (top) -- (btm);
\node[phase, below=5mm of btm, anchor=north west] (fine) {Phase 2};

\coordinate (top) at ($(top) + (35mm,0)$);
\coordinate (btm) at ($(btm) + (35mm,0)$);
\fill[fill=gray] (top |- threephaset1.north east) rectangle ($(btm |- threephaset1.south east) + (3*2.0mm, 0mm)$);
\coordinate (num) at (top |- threephaset1.north east);
\node[num, below right=0.5mm of num, xshift=8mm] () {20};
\fill[fill=gray] (top |- threephaset2.north east) rectangle ($(btm |- threephaset2.south east) + (3*7.0mm, 0mm)$);
\coordinate (num) at (top |- threephaset2.north east);
\node[num, white, below right=0.5mm of num, xshift=8mm] () {70};
\coordinate (num) at (top |- threephaset3.north east);
\node[num, below right=0.5mm of num, xshift=8mm] () {0};
\coordinate (num) at (top |- threephaset4.north east);
\node[num, below right=0.5mm of num, xshift=8mm] () {0};
\fill[fill=gray] (top |- threephaset5.north east) rectangle ($(btm |- threephaset5.south east) + (3*1.0mm, 0mm)$);
\coordinate (num) at (top |- threephaset5.north east);
\node[num, below right=0.5mm of num, xshift=8mm] () {10};
\draw[distrbar] (top) -- (btm);
\node[phase, below=5mm of btm, anchor=north west] (fine) {Phase 3};

\coordinate (timestart) at ($(pre.north west)+(0,2.5mm)$);
\coordinate (timeend)  at ($(fine.north east)+(-4mm,2.5mm)$);
\draw[pil] (timestart) -- (timeend);
\node[at=(timeend), anchor=west, xshift=1mm, phase] (time) {time};

\end{tikzpicture}%
}
  \caption{The task mix schedule used in the 3-phase scheduled multi-task learning experiment.
           The 2-phase schedule is the same, except it omits the third phase, continuing the second phase until the end of training.
  \label{fig:experimentschedules}}
\end{minipage}
\end{figure}

Figure~\ref{fig:learningcurve} shows the learning curves on the development set
and Table~\ref{tab:results_curriculum} the evaluations on the test set
for different configurations of transfer learning.

\emph{Multi-task without schedule} is trained with a constant task mixing distribution.
The result marked \emph{HRL pretraining, LRL fine-tuning} uses a mix of HRL translation and autoencoder tasks for pretraining,
and only a single task---LRL translation---for fine-tuning,
and is thus fully sequential in terms of languages.
It quickly overfits in the fine-tuning phase.

The models using scheduled multi-task learning combine sequential and parallel transfer.
In \emph{2-phase scheduled multi-task}, LRL tasks are not used in the pretraining phase, but a mix of tasks is used for fine-tuning.
It gives a benefit of +2.4 BLEU compared to the model fine-tuning on only LRL tasks,
and +1.77 BLEU compared to training with a constant mixing distribution.
The \emph{3-phase scheduled multi-task} adds a third phase training mostly on LRL tasks.
A small proportion of HRL translation is included to delay overfitting.
The model again overfits in the final phase, but does reach a higher score before doing so.
The 3-phase task mixing schedule is shown in Figure~\ref{fig:experimentschedules}.

\begin{table}[ht]
\centering
\caption{Results for scheduled multi-task learning. \label{tab:results_curriculum}}
\setlength{\tabcolsep}{0.3em}
\begin{tabular}{lllllllrrrrrrrrr}
\toprule
        &          &                     & \multicolumn{3}{c}{Autoencoder} & \multicolumn{3}{c}{\eng--\est} & \multicolumn{3}{c}{\eng--\dan} & \multicolumn{3}{c}{\eng--\slo} \\
\cmidrule(lr){4-6} \cmidrule(lr){7-9} \cmidrule(lr){10-12} \cmidrule(lr){13-15}
Method  & ML       & BT                            &             SRC & HRL & LRL &     chrF1 &  BLEU &  rare & chrF1 &  BLEU &  rare & chrF1 &  BLEU &  rare \\
\midrule
          3-phase scheduled multi-task &      \cm  &           & \cm & \cm & \cm &     51.71 & 13.94 & 33.96 & 50.1  & 13.7  & 54.6  & 50.1  & 14.1  & 69.7 \\
          2-phase scheduled multi-task &      \cm  &           & \cm & \cm & \cm &     51.42 & 13.75 & 33.83 & 49.8  & 13.5  & 55.3  & 50.2  & 14.0  & 69.7 \\
               Multi-task w/o schedule &      \cm  &           & \cm & \cm & \cm &     48.62 & 11.98 & 29.16 & 48.0  & 12.2  & 52.5  & 48.3  & 12.6  & 68.8 \\
      HRL pretraining, LRL fine-tuning &      \cm  &           & \cm & \cm & \cm &     48.15 & 11.35 & 29.88 & 47.8  & 11.6  & 49.9  & 47.0  & 11.4  & 66.3 \\
                Subnetwork pretraining &      \cm  &           & \cm & \cm & \cm &     47.74 & 11.17 & 26.93 \\

\bottomrule
\end{tabular}
\setlength{\tabcolsep}{0.5em}   
\end{table}


\citet{torrey2009transfer} describe three ways in which transfer learning can benefit training:
1) higher performance at the very beginning of learning, 2) steeper learning curve, and 3) higher asymptotic performance.
When pretraining the encoder and decoder on source and target autoencoder tasks respectively, we see the first of these, but not the other two:
for \eng--\est\, NMT training at first improves faster than with random initialization, but converges to a worse final model.
As the approach was clearly inferior, we did not use it for the other language pairs.
However, we have not tested pretraining on a next token prediction or masked language modeling task.

\subsubsection{Dataset augmentation -- Subword regularization}

Table~\ref{tab:results_subwreg} shows an improvement between +0.08 and +0.55 BLEU from using
subword regularization as the only noise model, without the use of an autoencoder.

\begin{table}[ht]
\centering
\caption{Results with subword regularization (SWR). \label{tab:results_subwreg}}
\setlength{\tabcolsep}{0.3em}
\begin{tabular}{lllllllrrrrrrrrr}
\toprule
        &          &                     & \multicolumn{3}{c}{Autoencoder} & \multicolumn{3}{c}{\eng--\est} & \multicolumn{3}{c}{\eng--\dan} & \multicolumn{3}{c}{\eng--\slo} \\
\cmidrule(lr){4-6} \cmidrule(lr){7-9} \cmidrule(lr){10-12} \cmidrule(lr){13-15}
Method  & ML       & BT                            &             SRC & HRL & LRL &    chrF1 &  BLEU &  rare & chrF1 &  BLEU &  rare & chrF1 &  BLEU &  rare \\
\midrule
                                   SWR &      \cm  &           &     &     &     &    50.09 & 12.90 & 33.20 & 49.57 & 13.13 & 54.21 & 49.83 & 13.97 & 68.79 \\
                                no SWR &      \cm  &           &     &     &     &    49.77 & 12.57 & 31.14 & 49.27 & 13.05 & 53.66 & 49.27 & 13.42 & 69.07 \\
\bottomrule
\end{tabular}
\setlength{\tabcolsep}{0.5em}   
\end{table}

\subsubsection{Dataset augmentation -- Autoencoder}

Table~\ref{tab:results_noises} shows an ablation experiment for the noise model.
When compared against only using the subword regularization,
the additional noises give between +0.2 and +0.5 BLEU.
All parts of the noise model are individually ablated:
the most important is local reordering, which when omitted causes a decrease of -0.36 BLEU.
The full noise model includes subword regularization.
When subword regularization is ablated, we turn it entirely off, both for the parallel data and the autoencoder.
Word boundary noise, taboo sampling, and insertions are not included in our full noise model,
as they did not show a benefit on the development set.
However, word boundary noise gives +0.2 BLEU
and taboo sampling +0.09 BLEU on the test set.

\begin{table}[ht]
\centering
\caption{Ablation results for noise model. Ordered by decreasing BLEU. \label{tab:results_noises}}
\begin{tabular}{lllllllrrr}
\toprule
        &          &                               & \multicolumn{3}{c}{Autoencoder} & \multicolumn{3}{c}{\eng--\est} \\
\cmidrule(lr){4-6} \cmidrule(lr){7-9}
Method  & ML       & BT                            &             SRC & HRL & LRL &  chrF-1.0 &  BLEU &  rare \\
\midrule
                 + Word boundary noise &      \cm  &           & \cm & \cm & \cm &     51.56 & 13.95 & 33.20 \\
                      + Taboo sampling &      \cm  &           & \cm & \cm & \cm &     51.23 & 13.84 & 33.81 \\
                               No drop &      \cm  &           & \cm & \cm & \cm &     51.48 & 13.79 & 33.89 \\
                            Full noise &      \cm  &           & \cm & \cm & \cm &     51.42 & 13.75 & 33.83 \\
                           + Insertion &      \cm  &           & \cm & \cm & \cm &     50.88 & 13.74 & 33.51 \\
                        Only switchout &      \cm  &           & \cm & \cm & \cm &     50.78 & 13.49 & 32.21 \\
                                No SWR &      \cm  &           & \cm & \cm & \cm &     50.71 & 13.46 & 32.18 \\
                              Only SWR &      \cm  &           & \cm & \cm & \cm &     50.96 & 13.43 & 32.85 \\
                            No reorder &      \cm  &           & \cm & \cm & \cm &     50.90 & 13.39 & 33.03 \\

\bottomrule
\end{tabular}
\end{table}

We also consider for which languages an autoencoder task should be added.
Table~\ref{tab:results_aelangs} shows variants starting from no autoencoder,
adding autoencoders one by one first for the low-resource target language,
then for the source language and finally for the high-resource target language.
The best combination uses source and LRL,
with the SRC autoencoder giving a gain of +0.11 BLEU over only using the LRL.
The HRL autoencoder is detrimental, and leaving it out gives +0.29 BLEU.

\begin{table}[ht]
\centering
\caption{Autoencoder language tasks. \label{tab:results_aelangs}}
\begin{tabular}{lllllllrrr}
\toprule
        &          &                               & \multicolumn{3}{c}{Autoencoder} & \multicolumn{3}{c}{\eng--\est} \\
\cmidrule(lr){4-6} \cmidrule(lr){7-9}
Method  & ML       & BT                            &             SRC & HRL & LRL &  chrF-1.0 &  BLEU &  rare \\
\midrule
                            SRC+LRL AE &      \cm  &           & \cm &     & \cm &     51.71 & 14.04 & 34.79 \\
                                LRL AE &      \cm  &           &     &     & \cm &     51.41 & 13.93 & 33.57 \\
                        SRC+HRL+LRL AE &      \cm  &           & \cm & \cm & \cm &     51.42 & 13.75 & 33.83 \\
                                 No AE &      \cm  &           &     &     &     &     50.09 & 12.90 & 33.20 \\

\bottomrule
\end{tabular}
\end{table}

\subsubsection{Dataset augmentation -- Back-translation}

Table~\ref{tab:results_bt} shows the improvements gained using back-translated synthetic data.
We weight the natural and synthetic LRL data equally.
Back-translation is generally effective, giving a benefit between +1.31 and +4.46 BLEU.
When using back-translated data, the autoencoder task is less effective,
with small improvements to Character F$_1$ but inconsistent results for the other measures.
Note that back-translation is not a silver bullet.
The \emph{Vanilla BT} system uses only back-translation, but not multilingual training or autoencoder:
the back-translation is performed with a weak model trained only on the low-resource parallel data,
and then a forward model is trained augmented only by this low-quality back-translation.
The performance when using only back-translation is very low: only +2.87 BLEU better than the vanilla model without back-translation.
The high-quality back-translation together with multilingual training gives an +12.7 BLEU increase over the vanilla back-translation.

\begin{table}[ht]
\centering
\caption{Results using back-translation.
         \vanillabt\, indicates the use of a low-quality back-translation made with a non-multilingual non-autoencoder vanilla BT model.
    \label{tab:results_bt}}
\setlength{\tabcolsep}{0.3em}
\begin{tabular}{lllllllrrrrrrrrr}
\toprule
        &          &                     & \multicolumn{3}{c}{Autoencoder} & \multicolumn{3}{c}{\eng--\est} & \multicolumn{3}{c}{\eng--\dan} & \multicolumn{3}{c}{\eng--\slo} \\
\cmidrule(lr){4-6} \cmidrule(lr){7-9} \cmidrule(lr){10-12} \cmidrule(lr){13-15}
Method  & ML       & BT                            &             SRC & HRL & LRL &    chrF1 &  BLEU &  rare & chrF1 &  BLEU &  rare & chrF1 &  BLEU &  rare \\
\midrule
                               Full BT &      \cm  &       \cm & \cm &     & \cm &    56.45 & 18.05 & 41.13 & 51.27 & 14.80 & 56.63 & 52.80 & 16.87 & 70.97 \\
                        No AE, full BT &      \cm  &       \cm &     &     &     &    56.33 & 18.15 & 40.85 & 51.20 & 15.00 & 57.39 & 52.65 & 16.63 & 70.82 \\
                             AE, no BT &      \cm  &           & \cm &     & \cm &    51.71 & 14.04 & 34.79 & 50.06 & 13.92 & 54.58 & 50.19 & 14.02 & 69.94 \\
                            Vanilla BT &           &\vanillabt &     &     &     &    36.12 &  5.51 & 13.25 \\
\bottomrule
\end{tabular}
\setlength{\tabcolsep}{0.5em}   
\end{table}


\subsubsection{Amount of low-resource language data}
\begin{figure}
  \centering
  \includegraphics[width=0.45\textwidth]{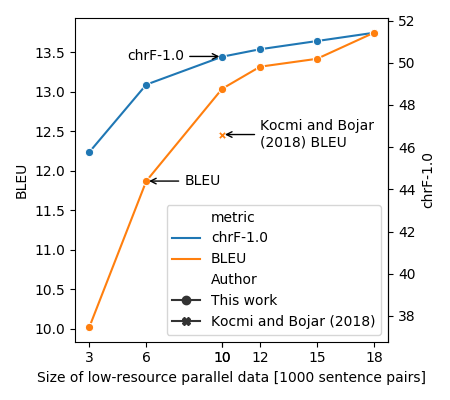}
  \caption{Varying the amount of low-resource data.
           Multilingual models, with SRC+HRL+LRL autoencoder and full noise model.
           Results on English$\rightarrow$Estonian newstest2018.}
  \label{fig:lowres}
\end{figure}

Figure~\ref{fig:lowres} shows how the performance degrades when the low-resource parallel data is reduced.
Each set is subsampled from the previous larger set.
All models use multilingual training with scheduled multi-task learning, and SRC+HRL+LRL autoencoders.
Down to 10k parallel sentences the performance stays reasonable, after which it rapidly deteriorates.

Also plotted is a 10k sentence pair baseline by \citet{kocmi2018trivial}, reaching 12.46 BLEU in a similar setting on the same test set.
Our result at 10k is 13.04 BLEU, or +0.68.

\subsubsection{Relatedness of the target languages}

Table~\ref{tab:results_cross} shows the results of using an unrelated but larger HRL (Czech).
The results favor transfer from the related HRL (Finnish), by +0.92 BLEU.
The difference in favor of the related HRL is largest for the rare words.

\begin{table}[ht]
\centering
\caption{HRL language relatedness. \label{tab:results_cross}}
\begin{tabular}{lllllllrrr}
\toprule
        &          &                               & \multicolumn{3}{c}{Autoencoder} & \multicolumn{3}{c}{\eng--\est} \\
\cmidrule(lr){4-6} \cmidrule(lr){7-9}
Method  & ML       & BT                            &             SRC & HRL & LRL &  chrF-1.0 &  BLEU &  rare \\
\midrule
                        Within family  &      \fin &           & \cm &     & \cm &     51.71 & 14.04 & 34.79 \\
                         Cross family  &      \cze &           & \cm &     & \cm &     50.20 & 13.12 & 30.69 \\

\bottomrule
\end{tabular}
\end{table}

Previously, \citet{zoph2016transfer} and \citet{dabre2017empirical} find that related parent languages result in better transfer.
However, \citet{kocmi2018trivial} find in the case of Estonian that a bigger parent (Czech) gave better results than a more related parent (Finnish).
Our results contradict \citet{kocmi2018trivial} and agree with the prior literature.

\subsubsection{Norwegian bokmål $\rightarrow$ Finnish + North Sámi}

We apply the findings of the previous experiments to the low-resource pair Norwegian bokmål to North Sámi.
We use a larger task mix weight for the LRL task (40 SRC-HRL / 30 SRC-LRL / 30 BT) to account for the larger LRL parallel data.
Table~\ref{tab:results_apertium} shows
the results to be similar to the results of the other languages,
with benefit from multilingual training, autoencoder task and back-translation.

\begin{table}[ht]
\centering
\caption{
    Results on Norwegian Bokmål--North Sámi Apertium story. \label{tab:results_apertium}}
\begin{tabular}{lllllllrrr}
\toprule
        &          &                               & \multicolumn{3}{c}{Autoencoder} &  \multicolumn{3}{c}{\nob--\sme} \\
\cmidrule(lr){4-6} \cmidrule(lr){7-9}
Method  & ML       & BT                            &          SRC & HRL & LRL &            chrF-1.0 &  BLEU &  rare \\
\midrule

                               ML, AE, BT &       \cm & \cm & \cm &     & \cm &               57.27 & 24.40 & 35.62 \\
                                   ML, AE &       \cm &     & \cm &     & \cm &               54.86 & 21.07 & 21.54 \\
                                  Vanilla &           &     &     &     &     &               45.97 & 15.64 & 21.05 \\
\bottomrule
\end{tabular}
\end{table}

\subsection{Discussion}

In our experiments for four asymmetric-resource one-to-many translation tasks,
we find that
the largest gains come from cross-lingual transfer (up to +12.7 BLEU),
back-translation (up to +4.46 BLEU),
and scheduled multi-task learning (up to +2.4 BLEU).
To sum up our findings related to the questions asked in the introduction:

On cross-lingual transfer,
we find that applying scheduled multi-task learning
is superior to both fully sequential and fully parallel transfer.
In scheduled multi-task learning,
the model is first pretrained on a mix of only high-resource tasks
and then fine-tuned using a mix of both high- and low-resource tasks.
A second fine-tuning phase only on the low-resource tasks is prone to overfitting.

On exploiting monolingual data,
a low-resource target-language autoencoder is beneficial, even when using multilingual training,
but inconclusive together with back-translation.
A source-language autoencoder is also helpful, to a lesser degree, but a high-resource target autoencoder is not.
A noise model including subword regularization, reordering, and deletion is beneficial.
The results for substitutions and the proposed taboo sampling method are inconclusive.

On vocabulary construction,
Morfessor EM+Prune is superior to SentencePiece in this translation setting, for a gain of +0.6 BLEU.
As the methods use the same training algorithm, it indicates that the prior used in Morfessor is beneficial in finding efficient subword lexicons.
The vocabulary size has less effect (up to 0.5 BLEU for sizes between 8k and 20k) on the results.
Subword lexicon size has been considered an important parameter to tune~\citep{sennrich2019revisiting,salesky2020optimizing}.
Also our preliminary experiments of low-resource NMT without subword regularization suggested a more substantial effect for the lexicon size.
It seems that the subword sampling procedure (and perhaps the autoencoder task)
lessens the impact of the subword vocabulary size.

Regarding available data and languages,
larger low-resource parallel data give better results, but diminishing returns are already reached after 10k sentences.
We find language relatedness to be more important than parent language size in highly asymmetrical transfer.
\citet{sennrich2019revisiting} find that smaller models and batch sizes work better in low-resource settings.
We find that large models are better whenever auxiliary multilingual or monolingual data is used.
While in the vanilla setting, the smaller model is better, it still falls far behind the models using additional data.

Among the translation tasks,
we get the lowest scores in the English--Danish translation.
While Danish has the smallest LRL monolingual corpus,
as the same order is observed also for the models not using monolingual data,
the reason must lie elsewhere, possibly in the difficulty of the JRC-Acquis corpus.
The autoencoder task has the largest benefit for English--Estonian.
In the Norwegian--North Sámi experiment the size of the low-resource parallel data
is an order of magnitude larger than in the other experiments, but the results remain similar.
Due to the small size of the test set, we include the entire translation output in 
Ancillary File \texttt{translated.apertium.story.txt}.

The three evaluation measures---BLEU, Character F$_1$, and rare words F$_1$---generally agree.
Some exceptions include
ablation of the subword regularization
and using SwitchOut as the sole noise model,
which hurt in particular the rare words more than BLEU.
Turning off the autoencoder has the least effect on rare words,
even giving a slight improvement for \eng--\dan\ when using back-translation.

Our results again underscore the need to gather parallel data for low-resource language pairs.
This may be possible to accomplish at reasonable cost, as 10k sentence pairs already goes a long way.
Monolingual corpora of high quality and quantity are also of great importance as auxiliary data for MT.

\section{Conclusion}

When training a neural translation model for low-resource languages with limited parallel training data,
it is important to make use of efficient methods for cross-lingual learning, data augmentation, and subword segmentation.
Our experiments in asymmetric-resourced one-to-many translation show that
the largest individual improvements come from any cross-lingual transfer learning
and augmenting the training data with back-translation.
However, considerable benefits are gained also by less common approaches:
scheduled multi-task learning, subword regularization, and a denoising autoencoder with multiple noise models.
For this reason, we strongly recommend that NMT frameworks should include a dataloader with the ability to
(a) sample noisy minibatches for training and
(b) use a schedule for controlling the mixing of different tasks.
Subword sampling requires a probabilistic segmentation model such as SentencePiece or Morfessor, making them preferable to the more common BPE method.
Both our data loader implementation for the OpenNMT-py system and the Morfessor EM+Prune software are available with non-restrictive licenses.

\begin{acknowledgements}
This study has been supported by the MeMAD project,
funded by the European Union's Horizon 2020
research and innovation programme~(grant agreement No~780069),
and the FoTran project, funded by the European Research Council~(ERC) under the European Union's Horizon 2020 research and innovation programme~(grant agreement No~771113).
Computer resources within the Aalto University School of Science ``Science-IT'' project were used.
\end{acknowledgements}

\bibliographystyle{spbasic}      
\bibliography{2020_specialissue}

\end{document}